\documentclass[11pt,letterpaper]{article}
\usepackage{amssymb,latexsym,amsmath,amsthm}
\usepackage{amsmath, amsfonts, amsthm,amssymb}
\usepackage{amsfonts, amsthm}
\usepackage{epsfig}
\usepackage{makeidx}
\usepackage{fullpage}
\usepackage{graphicx}
\theoremstyle{plain}

\newtheorem{theorem}{Theorem}
\newtheorem{lem}{Lemma}
\newtheorem{prop}{Proposition}

\newenvironment{carlist}
 {\begin{list}{$\bullet$}
 {\setlength{\topsep}{0in} \setlength{\partopsep}{0in}
  \setlength{\parsep}{0in} \setlength{\itemsep}{\parskip}
  \setlength{\leftmargin}{0.07in} \setlength{\rightmargin}{0.08in}
  \setlength{\listparindent}{0in} \setlength{\labelwidth}{0.08in}
  \setlength{\labelsep}{0.1in} \setlength{\itemindent}{0in}}}
 {\end{list}}

\newcommand{\bcar}{\begin{carlist}}
\newcommand{\ecar}{\end{carlist}}

\newenvironment{carliste}
 {\begin{list}x
 {\setlength{\topsep}{0in} \setlength{\partopsep}{0in}
  \setlength{\parsep}{0in} \setlength{\itemsep}{\parskip}
  \setlength{\leftmargin}{0.07in} \setlength{\rightmargin}{0.08in}
  \setlength{\listparindent}{0in} \setlength{\labelwidth}{0.08in}
  \setlength{\labelsep}{0.1in} \setlength{\itemindent}{0in}}}
 {\end{list}}

\newcommand{\bcare}{\begin{carliste}}
\newcommand{\ecare}{\end{carliste}}

\makeatletter
\long\def\@makecaption#1#2{
        \vskip 0.8ex
        \setbox\@tempboxa\hbox{\small {\bf #1:} #2}
        \parindent 1.5em  
        \dimen0=\hsize
        \advance\dimen0 by -3em
        \ifdim \wd\@tempboxa >\dimen0
                \hbox to \hsize{
                        \parindent 0em
                        \hfil 
                        \parbox{\dimen0}{\def\baselinestretch{0.96}\small
                                {\bf #1.} #2
                                } 
                        \hfil}
        \else \hbox to \hsize{\hfil \box\@tempboxa \hfil}
        \fi
        }
\makeatother

\long\def\comment#1{}

\makeatletter
\def\@cite#1#2{[\if@tempswa #2 \fi #1]}
\makeatother

\makeatletter
\long\def\barenote#1{
    \insert\footins{\footnotesize
    \interlinepenalty\interfootnotelinepenalty 
    \splittopskip\footnotesep
    \splitmaxdepth \dp\strutbox \floatingpenalty \@MM
    \hsize\columnwidth \@parboxrestore
    {\rule{\z@}{\footnotesep}\ignorespaces
      #1\strut}}}
\makeatother

\newcommand{\Ind}{\ensuremath{\mathbb{I\,}}}

\newcommand{\bit}{\begin{itemize}}
\newcommand{\eit}{\end{itemize}}
\newcommand{\ben}{\begin{enumerate}}
\newcommand{\een}{\end{enumerate}}

\newcommand{\bear}{\begin{eqnarray}}
\newcommand{\eear}{\end{eqnarray}}

\newcommand{\prob}{\ensuremath{{\mathbb{P}}}}

\newcommand{\order}{{\mathcal{O}}}

\newcommand{\ones}{\ensuremath{\mathbf{1}}}

\newcommand{\inv}[1]{\ensuremath{#1^{-1}}}

\newcommand{\Exs}{\ensuremath{{\mathbb{E}}}}

\newcommand{\beq}{\begin{quotation}}
\newcommand{\enq}{\end{quotation}}

\newcommand{\estart}{\begin{equation}}
\newcommand{\eend}{\end{equation}}

\newcommand{\widgraph}[2]{\includegraphics[keepaspectratio,width=#1]{#2}}

\newcommand{\edist}{\ensuremath{\overset{d}{=}}}

\newcommand{\defn}{\ensuremath{:  =}}

\newcommand{\Ysca}{{{Y}}}

\newcommand{\Wsca}{{{W}}}

\newcommand{\bec}{\begin{center}}
\newcommand{\enc}{\end{center}}

\newcommand{\beit}{\begin{itemize}}
\newcommand{\enit}{\end{itemize}}

\newcommand{\been}{\begin{enumerate}}
\newcommand{\enen}{\end{enumerate}}

\newcommand{\comsl}{\begin{slide}}
\newcommand{\comspor}{\begin{slide*}}

\newcommand{\comsld}[2]{\begin{slide}[#1,#2]}
\newcommand{\comspord}[2]{\begin{slide*}[#1,#2]}

\newcommand{\mendsl}{\end{slide}}
\newcommand{\mendspo}{\end{slide*}}

\newcommand{\estim}[1]{\ensuremath{\widehat{#1}}}

\newcommand{\wtil}[1]{\ensuremath{\widetilde{#1}}}

\newcommand{\real}{\ensuremath{{\mathbb{R}}}}

\DeclareMathOperator{\var}{var}

\DeclareMathOperator{\sign}{sign}

\newcommand{\Amat}{\ensuremath{X}}

\newcommand{\spgam}{\ensuremath{\gamma}}

\newcommand{\msparse}{\ensuremath{\gamma}}
\newcommand{\sval}{\ensuremath{s}}
\newcommand{\mprob}{\ensuremath{\mathbb{P}}}
\newcommand{\MGF}{\ensuremath{\mathbb{M}}}
\newcommand{\betastar}{\ensuremath{\beta^*}}
\newcommand{\betahat}{\ensuremath{\estim{\beta}}}

\newcommand{\arow}{\ensuremath{x}}
\newcommand{\myparagraph}[1]{\noindent {\bf{#1}}}

\newcommand{\PopCov}{\ensuremath{\Sigma}}
\newcommand{\SamCov}{\ensuremath{\widehat{\PopCov}}}
\newcommand{\Sset}{\ensuremath{{S}}}
\newcommand{\Ssetcomp}{\ensuremath{\Sset^c}}

\newcommand{\pdim}{\ensuremath{p}}
\newcommand{\numobs}{\ensuremath{n}}
\newcommand{\kdim}{\ensuremath{k}}
\newcommand{\minval}{\ensuremath{\beta}_{min}}
\newcommand{\regpar}{\ensuremath{\rho}}
\newcommand{\regparn}{\ensuremath{\regpar_\numobs}}

\newcommand{\Length}[2]{\ensuremath{\Vert\,#1\,\Vert_{#2}}}

\newcommand{\vershprobb}[1]{\exp(-{#1} \log(\pdim - \kdim))}

\newcommand{\LHS}[1]{\frac{C}{#1} \sqrt{\max\left \{
\frac{1}{\kdim{#1}}, \frac{\log {#1} \log(\pdim - \kdim)}{{#1}
\log(\pdim - \kdim)} \right \}}} 

\newcommand{\ellone}{\left(1- \mathcal{O}\Big(\frac{1}{\spgam}
\sqrt{\max\left \{ \frac{1}{\kdim}, \frac{\log \log(\pdim -
\kdim)}{\log(\pdim - \kdim)} \right \}}\Big) \right)^{-1}}
\newcommand{\elltwo}{\Biggr(1+ \frac{1}{2 \sqrt{\kdim} \spgam}\Biggr)
\; \Biggr[1+ \mathcal{O}\Big(\frac{1}{\spgam} \sqrt{\max\left \{
\frac{1}{\kdim(\spgam-\frac{1}{2\sqrt{\kdim}} )},
\frac{\log{(\spgam+\frac{1}{2\sqrt{\kdim}}) \log(\pdim - \kdim)}}{(
\spgam-\frac{1}{2\sqrt{\kdim}})\log(\pdim - \kdim)} \right \}}\Big)
\Biggr]}

\newcommand{\LHScortwo}{\mathcal{O}\huge(\frac{1}{\spgam} \sqrt{\max\left \{ \frac{\log{(\kdim)}}{\kdim\log{(\pdim - \kdim)}}, \frac{\log \log(\pdim - \kdim)}{\log(\pdim - \kdim)} \right \} }\huge)}

\newcommand{\mycomplement}{\ensuremath{c}}
\newcommand{\zhat}{\ensuremath{\estim{z}}}
\newcommand{\Sbar}{\ensuremath{{S^\mycomplement}}}

\newcommand{\contpar}{\ensuremath{\theta}}

\newcommand{\dualvec}{\ensuremath{\widehat{z}}}

\newcommand{\signvec}{\ensuremath{{\vec{v}}}}

\newcommand{\myones}{\ensuremath{\vec{1}}}

\newcommand{\Va}{\ensuremath{V^a}}
\newcommand{\Vb}{\ensuremath{V^b}}
\newcommand{\Uvar}{\ensuremath{U}}
\newcommand{\Vvar}{\ensuremath{V}}
\newcommand{\Event}{\ensuremath{\mathcal{E}}}

\newcommand{\keyvector}{\ensuremath{h}}

\newcommand{\Hvar}{\ensuremath{H}}
\newcommand{\Zvar}{\ensuremath{Z}}

\newcommand{\Randmat}{\ensuremath{\Gamma}}

\newcommand{\Ber}{\ensuremath{\operatorname{Ber}}}

\newcommand{\inverone}{\ensuremath{t}}
\newcommand{\invertwo}{\ensuremath{\theta}}

\newcommand{\Normal}{\ensuremath{N}}

\newcommand{\fone}{\ensuremath{f_1}}
\newcommand{\ftwo}{\ensuremath{f_2}}

\newcommand{\Spectral}[1]{\ensuremath{|\!|\!| #1 | \! | \!|_2}}
\newcommand{\myinv}[1]{\ensuremath{{#1}^{-1}}}

\newcommand{\Tail}{\ensuremath{\mathcal{T}}}
\newcommand{\Tailtwo}{\ensuremath{\mathcal{T}_2}}

\newcommand{\Projorth}[1]{\ensuremath{\Pi^{\perp}_{#1}}}
\newcommand{\umean}{\ensuremath{m}}
\newcommand{\uvariance}{\ensuremath{\psi}}

\newcommand{\Bin}{\ensuremath{\operatorname{Bin}}}

\newlength{\widebarargwidth}
\newlength{\widebarargheight}
\newlength{\widebarargdepth}
\DeclareRobustCommand{\widebar}[1]{%
  \settowidth{\widebarargwidth}{\ensuremath{#1}}%
  \settoheight{\widebarargheight}{\ensuremath{#1}}%
  \settodepth{\widebarargdepth}{\ensuremath{#1}}%
  \addtolength{\widebarargwidth}{-0.3\widebarargheight}%
  \addtolength{\widebarargwidth}{-0.3\widebarargdepth}%
  \makebox[0pt][l]{\hspace{0.3\widebarargheight}%
    \hspace{0.3\widebarargdepth}%
    \addtolength{\widebarargheight}{0.3ex}%
    \rule[\widebarargheight]{0.95\widebarargwidth}{0.1ex}}%
  {#1}}

\newcommand{\Ybar}{\ensuremath{\widebar{Y}}}

\begin{document}

\begin{center}

{\bf{\LARGE{High-dimensional subset recovery in noise: Sparsified
measurements without loss of statistical efficiency}}}

\vspace*{.2in}

{\large{
\begin{tabular}{ccc}
Dapo Omidiran$^\star$ & &  Martin J. Wainwright$^{\dagger,\star}$ \\
\end{tabular}
}}

\vspace*{.2in}

\begin{tabular}{c}
Department of Statistics$^\dagger$, and \\
Department of Electrical Engineering and Computer Sciences$^\star$ \\
UC Berkeley,  Berkeley, CA  94720
\end{tabular}

\vspace*{.2in}

\today

\vspace*{.2in}

\begin{tabular}{c}
Technical Report, \\
Department of Statistics,  UC Berkeley
\end{tabular}

\end{center}


\begin{abstract}
We consider the problem of estimating the support of a vector $\beta^*
\in \mathbb{R}^{p}$ based on observations contaminated by noise.  A
significant body of work has studied behavior of $\ell_1$-relaxations
when applied to measurement matrices drawn from standard dense
ensembles (e.g., Gaussian, Bernoulli).  In this paper, we analyze
\emph{sparsified} measurement ensembles, and consider the trade-off
between measurement sparsity, as measured by the fraction $\gamma$ of
non-zero entries, and the statistical efficiency, as measured by the
minimal number of observations $n$ required for exact support recovery
with probability converging to one.  Our main result is to prove that
it is possible to let $\gamma \rightarrow 0$ at some rate, yielding
measurement matrices with a vanishing fraction of non-zeros per row
while retaining the same statistical efficiency as dense ensembles.  A
variety of simulation results confirm the sharpness of our theoretical
predictions.
\end{abstract}


\section{Introduction}

Recent years have witnessed a flurry of research on the recovery of
high-dimensional sparse signals (e.g., compressed
sensing~\cite{CandesTao05,Donoho04a,Tropp06}, graphical model
selection~\cite{Meinshausen06,RavWaiLaf08}, and sparse
approximation~\cite{Tropp06}).  In all of these settings, the basic
problem is to recover information about a high-dimensional signal
$\betastar \in \real^{\pdim}$, based on a set of $\numobs$
observations.  The signal $\betastar$ is assumed \emph{a priori} to be
sparse: either exactly $\kdim$-sparse, or lying within some
$\ell_q$-ball with $q < 1$.  A large body of theory has focused on the
behavior of various $\ell_1$-relaxations when applied to measurement
matrices drawn from the standard Gaussian
ensemble~\cite{Donoho04a,CandesTao05}, or more general random
ensembles satisfiying mutual incoherence
conditions~\cite{Meinshausen06,Wainwright06a}.

These standard random ensembles are dense, in that the number of
non-zero entries per measurement vector is of the same order as the
ambient signal dimension.  Such dense measurement matrices are
undesirable for practical applications (e.g., sensor networks), in
which it would be preferable to take measurements based on sparse
inner products. Sparse measurement matrices require significantly less
storage space, and have the potential for reduced algorithmic
complexity for signal recovery, since many algorithms for linear
programming, and conic programming more generally~\cite{Boyd02}, can
be accelerated by exploiting problem structure.  With this motivation,
a body of past work (e.g.~\cite{ComMut05,Gil06,SaBaBa06a,
XuHassibi07}), motivated by group testing or coding perspectives, has
studied compressed sensing methods based on sparse measurement
ensembles.  However, this body of work has focused on the case of
noiseless observations.

In contrast, this paper focuses on observations contaminated by
additive noise which, as we show, exhibits fundamentally different
behavior than the noiseless case. Our interest is not on sparse
measurement ensembles alone, but rather in understanding the
\emph{trade-off} between the degree of measurement sparsity, and its
statistical efficiency.  We assess measurement sparsity in terms of
the fraction $\spgam$ of non-zero entries in any particular row of the
measurement matrix, and we define statistical efficiency in terms of
the minimal number of measurements $\numobs$ required to recover the
correct support with probability converging to one.  Our interest can
be viewed in terms of experimental design: more precisely we ask: what
degree of measurement sparsity can be permitted without any compromise
in the statistical efficiency?  To bring sharp focus to the issue, we
analyze this question for exact subset recovery using
$\ell_1$-constrained quadratic programming, also known as the Lasso in
the statistics literature~\cite{Chen98,Tibshirani96}, where past work
on dense Gaussian measurement ensembles~\cite{Wainwright06a} provides
a precise characterization of its success/failure.  We characterize
the density of our measurement ensembles with a positive parameter
$\spgam \in (0,1]$, corresponding to the fraction of non-zero entries
per row. We first show that for all fixed $\spgam \in (0,1]$, the
statistical efficiency of the Lasso remains the same as with dense
measurement matrices.  We then prove that it is possible to let
$\spgam \rightarrow 0$ at some rate, as a function of the sample size
$\numobs$, signal length $\pdim$ and signal sparsity $\kdim$, yielding
measurement matrices with a vanishing fraction of non-zeroes per row
while requiring exactly the same number of observations as dense
measurement ensembles.  In general, in contrast to the noiseless
setting~\cite{XuHassibi07}, our theory still requires that the average
number of non-zeroes per column of the measurement matrix (i.e.,
$\spgam \numobs$) tend to infinity; however, under the loss function
considered here (exact signed support recovery), we prove that no
method can succeed with probability one if this condition does not
hold.  The remainder of this paper is organized as follows.  In
Section~\ref{SecMain}, we set up the problem more precisely, state our
main result, and discuss some of its implications.  In
Section~\ref{SecProofs}, we provide a high-level outline of the proof.

Work in this paper was presented in part at the International
Symposium on Information Theory in Toronto, Canada (July, 2008).  We
note that in concurrent and complementary work, Wang et
al.~\cite{WanWaiRam08} have analyzed the information-theoretic
limitations of sparse measurement matrices for exact support recovery.

\paragraph{Notation:} Throughout this paper, we use the following standard 
asymptotic notation: $f(n) = \mathcal{O}(g(n))$ if $f(n) \leq C g(n)$
for some constant $C < +\infty $; $f(n) = \Omega(g(n))$ if $f(n) \geq
c g(n)$ for some constant $c > 0$; and $f(n) = \Theta(g(n))$ if $f(n)
= \mathcal{O}(g(n))$ and $f(n) = \Omega(g(n))$.

\section{Problem set-up and main result}
\label{SecMain}

We begin by setting up the problem, stating our main result, and
discussing some of their consequences.

\subsection{Problem formulation}
\label{ProbFormulation}
Let $\betastar \in \real^\pdim$ be a fixed but unknown vector, with at
most $\kdim$ non-zero entries ($\kdim \leq \frac{\pdim}{2}$), and define its \emph{support set}
\begin{eqnarray}
\label{EqnDefnSset}
\Sset & \defn & \{ i \in \{1, \ldots, \pdim \} \, \mid \; \betastar_i
\neq 0 \}.
\end{eqnarray}
We use $\minval$ to denote the minimum value of $|\betastar|$ on its
support---that is, $\minval \defn \min_{i \in \Sset} |\beta^*_i|$.

Suppose that we make a set $\{\Ysca_1, \ldots, \Ysca_\numobs \}$ of
$\numobs$ independent and identically distributed (i.i.d.)
observations of the unknown vector $\betastar$, each of the form
\begin{eqnarray}
\label{EqnObsModel}
\Ysca_i & \defn & \arow_i^T \betastar + \Wsca_i,
\end{eqnarray}
where $\Wsca \sim \mathcal{N}(0, \sigma^2)$ is observation noise, and
$\arow_i \in \real^\pdim$ is a measurement vector.  It is convenient
to use $\Ysca =
\begin{bmatrix} \Ysca_1 & \Ysca_2 & \ldots & \Ysca_\numobs \end{bmatrix}^T$ to
denote the $\numobs$-vector of measurements, with similar notation for
the noise vector $\Wsca \in \real^\numobs$, and
\begin{eqnarray}
\Amat & = & \begin{bmatrix} \arow_1^T \\ \arow_2^T \\ \vdots \\
  \arow_\numobs^T \end{bmatrix}\; = \; \begin{bmatrix} \Amat_1 &
  \Amat_2 & \ldots & \Amat_\pdim \end{bmatrix}.
\end{eqnarray}
to denote the $\numobs \times \pdim$ measurement matrix.  With this
notation, the observation model can be written compactly as $\Ysca =
\Amat \betastar + \Wsca$.

Given some estimate $\betahat$, its error relative to the true
$\betastar$ can be assessed in various ways, depending on the
underlying application of interest.  For applications in compressed
sensing, various types of $\ell_q$ norms (i.e., \mbox{$\Exs \|\betahat
- \betastar\|_q$}) are well-motivated, whereas for statistical
prediction, it is most natural to study a predictive loss (e.g., $\Exs
\|\Amat \betahat - \Amat \betastar\|$).  For reasons of scientific
interpretation or for model selection purposes, the object of primary
interest is the support $\Sset$ of $\betastar$.  In this paper, we
consider a slightly stronger notion of model selection: in particular,
our goal is to recover the \emph{signed support} of the unknown
$\betastar$, as defined by the $\pdim$-vector $\Sset(\betastar)$ with
elements
\begin{eqnarray*}
\label{EqnDefnSignedSupp}
[{\Sset}(\betastar)]_i & \defn & \begin{cases} \sign(\betastar_i) &
\mbox{if $\betastar_i \neq 0$} \\
0 & \mbox{otherwise.}
			     \end{cases}
\end{eqnarray*}
Given some estimate $\betahat$, we study the probability
$\mprob[{\Sset}(\betahat) = {\Sset}(\betastar)]$ that it correctly
specifies the signed support.

The estimator that we analyze is $\ell_1$-constrained quadratic
programming (QP), also known as the Lasso~\cite{Tibshirani96} in the
statistics literature.  The Lasso generates an estimate $\betahat$ by
solving the regularized QP
\begin{eqnarray}
\label{EqnDefnLasso}
\betahat & = & \arg \min_{\beta \in \real^\pdim} \left \{ \frac{1}{2
\numobs} \| \Ysca - \Amat \beta \|_2^2 + \regparn \|\beta\|_1 \right
\},
\end{eqnarray}
where $\regparn > 0$ is a user-defined regularization parameter.  A
large body of past work has focused on the behavior of the Lasso for
both deterministic and random measurement matrices
(e.g.,~\cite{Donoho04b,Meinshausen06,Tropp06,Wainwright06a}).  Most
relevant here is the sharp threshold~\cite{Wainwright06a}
characterizing the success/failure of the Lasso when applied to
measurement matrices $\Amat$ drawn randomly from the standard Gaussian
ensemble (i.e., each element $\Amat_{ij} \sim \mathcal{N}(0,1)$
i.i.d.).  In particular, the Lasso undergoes a sharp threshold as a
function of the control parameter
\begin{eqnarray}
\label{EqnDefnThresh}
\contpar(\numobs, \pdim, \kdim) & \defn & \frac{\numobs}{2 \kdim \log(\pdim
  - \kdim)}.
\end{eqnarray}
For the standard Gaussian ensemble and sequences $(\numobs, \pdim,
\kdim)$ such that $\contpar(\numobs, \pdim, \kdim) > 1$, the probability
of Lasso success goes to one, whereas it converges to zero for sequences
for which $\contpar(\numobs, \pdim, \kdim) < 1$.  The main contribution
of this paper is to show that the same sharp threshold holds for
$\spgam$-sparsified measurement ensembles, including a subset for
which $\spgam \rightarrow 0$, so that each row of the measurement
matrix has a vanishing fraction of non-zero entries.

\subsection{Statement of main result}

A measurement matrix $\Amat \in \real^{\numobs \times \pdim}$ drawn
randomly from a Gaussian ensemble is dense, in that each row has
$\Theta(\pdim)$ non-zero entries.  The main focus of this paper is the
observation model~\eqref{EqnObsModel}, using measurement ensembles
that are designed to be sparse.  To formalize the notion of sparsity,
we let $\spgam \in (0,1]$ represent a \emph{measurement sparsity
parameter}, corresponding to the (average) fraction of non-zero
entries per row.  Our analysis allows the sparsity parameter
$\spgam(\numobs, \pdim, \kdim)$ to be a function of the triple
$(\numobs, \pdim, \kdim)$, but we typically suppress this explicit
dependence so as to simplify notation.  For a given choice of
$\spgam$, we consider measurement matrices $\Amat$ with i.i.d. entries
of the form
\begin{eqnarray}
\label{EqnSparseMeas}
\Amat_{ij} & \stackrel{d}{=} & \begin{cases} Z \sim \mathcal{N}(0, 1)
& \mbox{with probability $\spgam$} \\ 0 & \mbox{with probability
$1-\spgam$}.
				\end{cases}
\end{eqnarray}
By construction, the expected number of non-zero entries in each row
of $\Amat$ is $\spgam\pdim$.  It is straightforward to verify that for
any constant setting of $\spgam$, elements $\Amat_{ij}$ from the
ensemble~\eqref{EqnSparseMeas} are sub-Gaussian.  (A zero-mean random
variable $Z$ is sub-Gaussian~\cite{VershNew} if there exists some
constant $C > 0$ such that $\mprob[|Z| > t] \leq 2 \exp(-C t^2)$ for
all $t > 0$.)  For this reason, one would expect such ensembles to
obey similar scaling behavior as Gaussian ensembles, although possibly
with different constants.  In fact, the analysis of this paper
establishes exactly the same control parameter
threshold~\eqref{EqnDefnThresh} for $\spgam$-sparsified measurement
ensembles, for any fixed $\spgam \in (0,1)$, as the completely dense
case ($\spgam = 1$).  On the other hand, if $\spgam$ is allowed to
tend to zero, elements of the measurement matrix are no longer
sub-Gaussian with any fixed constant, since the variance of the
Gaussian mixture component scales non-trivially.  Nonetheless, our
analysis shows that for $\spgam \rightarrow 0$ suitably slowly, it is
possible to achieve the same statistical efficiency as the dense case.

In particular, we state the following result on conditions under which
the Lasso applied to sparsified ensembles has the same \emph{sample
complexity} as when applied to the dense (standard Gaussian) ensemble:\\
%
\newcommand{\thmeps}{\ensuremath{\epsilon}}
\begin{theorem}
\label{ThmMain}
Suppose that the measurement matrix $\Amat \in \real^{\numobs \times
\pdim}$ is drawn with i.i.d. entries according to the
\mbox{$\spgam$-sparsified} distribution~\eqref{EqnSparseMeas}.  Then
for any $\thmeps > 0$, if the sample size satisfies
\begin{eqnarray}
\label{EqnSampleSize}
\numobs & > & (2 + \thmeps) \kdim \log(\pdim - \kdim),
\end{eqnarray}
then the Lasso succeeds with probability one as $(\numobs, \pdim,
\kdim) \rightarrow +\infty$ in recovering the correct signed support
as long as
\begin{subequations}
\begin{eqnarray}
\label{mt11}
\frac{\numobs \regparn^2 \msparse}{\log{(\pdim-\kdim)}} & \rightarrow
& \infty \\
\label{mt3} 
\frac{\regparn}{\minval} \left(1+{\frac{\sqrt{\kdim}}{\msparse} \sqrt{
\frac{\log \log(\pdim-\kdim)}{\log(\pdim -\kdim)} }}\right) & \rightarrow& 0
\\
\label{mt00}
\msparse^3 {\min\left \{\kdim, \frac{\log(\pdim
    -\kdim)}{\log{\log(\pdim-\kdim)}} \right \}} &\rightarrow& \infty.
\end{eqnarray}
\end{subequations}
\end{theorem}

\myparagraph{Remarks:} 

\noindent (a) To provide intuition for Theorem~\ref{ThmMain}, it is
helpful to consider various special cases of the sparsity parameter
$\spgam$.  First, if $\spgam$ is a constant fixed to some value in
$(0,1]$, then it plays no role in the scaling, and
condition~\eqref{mt00} is always satisfied.  Furthermore,
condition~\eqref{mt11} is then the exact same as that of from previous
work~\cite{Wainwright06a} on dense measurement ensembles ($\spgam =
1$). However, condition~\eqref{mt3} is slightly weaker than the
corresponding condition from~\cite{Wainwright06a} in that $\minval$
must approach zero more slowly.  Depending on the exact behavior of
$\minval$, choosing $\regparn^2$ to decay slightly more slowly than
$\log \pdim/\numobs$ is sufficient to guarantee exact recovery with
$\numobs = \Theta(\kdim \log(\pdim - \kdim))$, meaning that we recover
exactly the same statistical efficiency as the dense case ($\spgam =
1$) for all constant measurement sparsities $\spgam \in (0,1)$.  At
least initially, one might think that reducing $\spgam$ should
increase the required number of observations, since it effectively
reduces the signal-to-noise ratio by a factor of $\spgam$.  However,
under high-dimensional scaling ($\pdim \rightarrow +\infty)$, the
dominant effect limiting the Lasso performance is the number ($\pdim -
\kdim$) of irrelevant factors, as opposed to the signal-to-noise ratio
(scaling of the minimum).  \\

\noindent (b) However, Theorem~\ref{ThmMain} also allows for general
scalings of the measurement sparsity $\spgam$ along with the triplet
$(\numobs, \pdim, \kdim)$.  More concretely, let us suppose for
simplicity that $\minval = \Theta(1)$.  Then over a range of signal
sparsities---say $\kdim = \alpha \pdim$, $\kdim =
\Theta(\sqrt{\pdim})$ or $\kdim = \Theta(\log(\pdim-\kdim))$,
corresponding respectively to linear sparsity, polynomial sparsity,
and exponential sparsity----we can choose a decaying measurement
sparsity, for instance
\begin{eqnarray}
\label{EqnSpecChoice} 
\spgam & = & \left[\frac{\log{\log{(\pdim -
\kdim)}}}{\log{(\pdim-\kdim)}}\right]^{\frac{1}{6}} \rightarrow 0
\end{eqnarray}
along with the regularization parameter
\mbox{$\regparn^2=\frac{\log{(\pdim-\kdim)}}{\numobs}
\sqrt{\frac{\log{(\pdim - \kdim)}}{\log{\log{(\pdim-\kdim)}}}}$} while
maintaining the same sample complexity (required number of
observations for support recovery) as the Lasso with dense measurement
matrices. \\


\noindent (c) Of course, the conditions of Theorem~\ref{ThmMain} do
not allow the measurement sparsity $\spgam$ to approach zero
arbitrarily quickly.  Rather, for any $\spgam$ guaranteeing exact
recovery, condition~\eqref{mt11} implies that the average number of
non-zero entries per column of $\Amat$ (namely, $\spgam \numobs$) must
tend to infinity.  (Indeed, with $\numobs = \Omega(\kdim \log(\pdim -
\kdim))$, our specific choice~\eqref{EqnSpecChoice} certainly
satisfies this constraint.)  A natural question is whether exact
recovery is possible using measurement matrices, either randomly drawn
or deterministically designed, with the average number of non-zeros
per row (namely $\spgam \numobs$) remaining bounded. In fact, under
the criterion of exactly recovering the signed
support~\eqref{EqnDefnSignedSupp}, no method can succeed with w.p. one
if $\spgam \numobs \minval^2$ remains bounded.

\begin{prop} 
\label{PropNecessary}
If $\spgam \numobs  \minval^2$ does not tend to infinity, then no
method can recover the signed support with probability one.
\end{prop}
%


\begin{proof}
We construct a sub-problem that must be solvable by any method capable
of performing exact signed support recovery.  Suppose that
$\betastar_1 = \minval \neq 0$ and that the column $\Amat_1$ has $n_1$
non-zero entries, say without loss of generality indices $i = 1,
\ldots, n_1$.  Now consider the problem of recovering the sign of
$\betastar_1$.  Let us extract the observations $i=1, \ldots, n_1$
that explicitly involve $\betastar_1$, writing
\begin{eqnarray*}
\label{EqnNewObs}
\Ysca_i & = & \Amat_{i1} \betastar_1 + \sum_{j \in T(i)} \Amat_{ij}
\betastar_j + \Wsca_i, \qquad i = 1, \ldots, n_1
\end{eqnarray*}
where $T(i)$ denotes the set of indices in row $i$ for which $\Amat_{ij}$ is non-zero, excluding index $1$.  Even assuming
that $\{ \betastar_j, j \in T(i) \}$ were perfectly known, this
observation model~\eqref{EqnNewObs} is at best equivalent to observing
$\betastar_1$ contaminated by constant variance additive Gaussian
noise, and our task is to distinguish whether $\betastar_1 = \minval$
or $\betastar_1 = -\minval$.  The average $\Ybar = \frac{1}{n_1}
\sum_{i=1}^{n_1} [\Ysca_i - \sum_{j \in T(i)} \Amat_{ij}
\betastar_j]$ is a sufficient statistic, following the
distribution $\Ybar \sim \Normal(\minval, \frac{\sigma^2}{n_1})$.  Unless the
effective signal-to-noise ratio, which is of the order $n_1
\minval^2$, goes to infinity, there will always be a constant
probability of error in distinguishing $\betastar_1 = \minval$ from
$\betastar_1 = -\minval$.  Under the $\spgam$-sparsified random
ensemble, we have $n_1 \leq (1+o(1)) \, \spgam \numobs$ with high probability,
so that no method can succeed unless $\spgam \numobs \minval^2$ goes
to infinity, as claimed.
\end{proof}
Note that the conditions in Theorem~\ref{ThmMain} imply that $\numobs
\spgam \minval^2 \rightarrow +\infty$.  In particular,
condition~\eqref{mt3} implies that $\regparn^2 = o(\minval^2)$, and
condition~\eqref{mt11} implies that $\numobs \spgam \regparn^2
\rightarrow +\infty$, which implies the condition of
Proposition~\ref{PropNecessary}.

\section{Proof of Theorem~\ref{ThmMain}}
\label{SecProofs}

This section is devoted to the proof of Theorem~\ref{ThmMain}.  We
begin with a high-level outline of the proof; as with previous work on
dense Gaussian ensembles~\cite{Wainwright06a}, the key is the notion
of a \emph{primal-dual witness} for exact signed support recovery. We
then proceed with the proof, divided into a sequence of separate
lemmas.  Analysis of ``sparsified'' matrices require results on
spectral properties of random matrices not covered by the standard
literature. The proofs of some of the more technical results are
deferred to the appendices.

\subsection{High-level overview of proof}
\label{high_level}

For the purposes of our proof, it is convenient to consider matrices
$\Amat \in \mathbb{R}^{\numobs \times \pdim}$ with i.i.d. entries of
the form
\begin{eqnarray}
\label{EqnRescaledMeas}
\Amat_{ij} & \stackrel{d}{=} & \begin{cases} Z \sim \mathcal{N}(0,
\frac{1}{\spgam}) & \mbox{with probability $\spgam$} \\ 0 & \mbox{with
probability $1-\spgam$}.
				\end{cases}
\end{eqnarray}
So as to obtain an equivalent observation model, we also reset the
variance of $\Wsca_i$ of each noise term $\Wsca_i$ to be
$\frac{\sigma^2}{\gamma}$.    Finally, we can assume without loss of
generality that $\sign(\beta^*_\Sset) = \myones \in \real^\kdim$.

Define the \emph{sample covariance matrix}
\begin{eqnarray}
\SamCov & \defn & \frac{1}{\numobs} \Amat^T \Amat \; = \;
\frac{1}{\numobs} \sum_{i=1}^\numobs \arow_i \arow_i^T.
\end{eqnarray}
Of particular importance to our analysis is the $\kdim \times \kdim$
sub-matrix $\SamCov_{\Sset \Sset}$.  For future reference, we
state the following claim, proved in Appendix~\ref{AppLemInvertible}:
\begin{lem}
\label{LemInvertible}
Under the conditions of Theorem~\ref{ThmMain}, the submatrix
 $\SamCov_{\Sset \Sset}$ is invertible with probability greater than
 $1- \mathcal{O}(\frac{1}{(\pdim - \kdim)^2})$.
\end{lem}

 The foundation of our proof is the following lemma: it provides
sufficient conditions for the Lasso~\eqref{EqnDefnLasso} to recover
the signed support set.
\begin{lem}[Primal-dual conditions for support recovery]  
\label{LemUnique}
Suppose that $\SamCov_{\Sset \Sset} \succ 0$, and that we can find a
primal vector $\betahat \in \real^\pdim$, and a subgradient vector
$\zhat \in \real^\pdim$ that satisfy the \emph{zero-subgradient
condition}
\begin{eqnarray}
\label{EqnZeroSub}
\SamCov \left(\betastar - \betahat \right) + \frac{1}{\numobs} \Amat^T
\Wsca + \regpar_\numobs \zhat & = & 0,
\end{eqnarray}
and the \emph{signed-support-recovery conditions}
\begin{subequations}
\label{EqnUnique}
\begin{eqnarray}
\label{EqnUniqueA}
\zhat_i & = & \sign(\beta^*_i) \qquad \mbox{for all $i \in \Sset$,} \\
\label{EqnUniqueB}
\estim{\beta}_j & = & 0  \qquad \mbox{for all $j \in \Sbar$,} \\
\label{EqnUniqueC}
|\zhat_j| & < & 1  \qquad \mbox{for all $j \in \Sbar$, and} \\
\label{EqnUniqueD}
\sign(\estim{\beta}_i) & = & \sign(\beta^*_i) \qquad \mbox{for all $i
\in \Sset$}.
\end{eqnarray}
\end{subequations}
Then $\widehat{\beta}$ is the unique optimal solution to the
Lasso~\eqref{EqnDefnLasso}, and recovers the correct signed support.
\end{lem}
\noindent See Appendix~\ref{AppLemUnique} for the proof of this claim.

Thus, given Lemmas~\ref{LemInvertible} and~\ref{LemUnique}, it
suffices to show that under the specified scaling of $(\numobs, \pdim,
\kdim)$, there exists a primal-dual pair $(\estim{\beta}, \dualvec)$
satisfying the conditions of Lemma~\ref{LemUnique}.  We establish the
existence of such a pair with the following constructive procedure:
\begin{enumerate}
\item[(a)]  We begin by setting $\estim{\beta}_{\Sbar} = 0$, and
$\dualvec_{\Sset} = \sign(\beta^*_\Sset)$.
\item[(b)]  Next we determine $\estim{\beta}_{\Sset}$ by solving
the linear system
\begin{eqnarray}
\SamCov_{\Sset \Sset} \left(\beta^*_\Sset - \estim{\beta}_\Sset
\right) + \frac{1}{\numobs} \Amat^T_\Sset \Wsca + \regparn
\sign(\beta^*_\Sset) & = & 0.
\end{eqnarray}
\item[(c)]  Finally, we determine $\dualvec_{\Sbar}$ by solving the
linear system:
\begin{eqnarray}
-\regparn \dualvec_{\Sbar} & = & \SamCov_{\Sbar \Sset}
 \left(\beta^*_\Sset - \estim{\beta}_{\Sset} \right) +
 \frac{1}{\numobs} \Amat^T_{\Sbar} \Wsca.
\end{eqnarray}
\end{enumerate}
By construction, this procedure satisfies the zero sub-gradient
condition~\eqref{EqnZeroSub}, as well as auxiliary
conditions~\eqref{EqnUniqueA} and~\eqref{EqnUniqueB}; it remains to
verify conditions~\eqref{EqnUniqueC} and~\eqref{EqnUniqueD}.

In order to complete these final two steps, it is helpful to
define the following random variables:
\begin{subequations}
\label{dual_variables}
\begin{eqnarray}
\label{EqnDefnVa}
\Va_j & \defn & \frac{1}{\numobs} \Amat^T_j \left\{ \Amat_{\Sset}
(\SamCov_{\Sset \Sset})^{-1} \myones \right \} \regparn \\
\label{EqnDefnVb}
\Vb_j & \defn & {\Amat_j}^{T} \left[\frac{1}{\numobs} \Amat_{\Sset}
(\SamCov_{\Sset \Sset})^{-1} X_{\Sset}^{T} - I_{\numobs \times
\numobs} \right] \frac{\Wsca}{\numobs}, \\
\label{EqnDefnUvar}
\Uvar_i & \defn & e_i^T \left( \SamCov_{\Sset \Sset}
\right)^{-1}\left[\frac{1}{\numobs} \Amat_\Sset ^T \Wsca - \regparn
\myones \right],
\end{eqnarray}
\end{subequations}
where $e_i \in \real^\kdim$ is the unit vector with one in position
$i$, and $\ones \in \real^\kdim$ is the all-ones vector.

A little bit of algebra (see Appendix~\ref{SecVUvar} for details)
shows that $\regparn \dualvec_j = \Va_j + \Vb_j$, and that $\Uvar_i =
\estim{\beta}_i - \beta^*_i$.  
Consequently, if we define the events
\begin{subequations}
\begin{eqnarray}
\label{nas1}
\Event(\Vvar) & \defn & \left\{ \max_{j \in
{\Ssetcomp}}|\Va_j + \Vb_j| < \regparn \right\} \\
\label{nas2} 
\Event(\Uvar) &\defn& \left\{ \max_{i \in \Sset} |\Uvar_i| \leq
\minval \right\},
\end{eqnarray}
\end{subequations}
where the minimum value $\minval$ was defined
previously as the minimum value of $|\betastar|$ on its support, then in order to establish that the
Lasso succeeds in recovering the exact signed support, it suffices to
show that $\mprob[\Event(\Vvar) \cap \Event(\Uvar)] \rightarrow 1$,

We decompose the proof of this final claim in the following three
lemmas.  As in the statement of Theorem~\ref{ThmMain}, suppose that
$\numobs > (2+\epsilon) \kdim \log(\pdim - \kdim)$, for some fixed
$\epsilon > 0$. 
\begin{lem}[Control of $\Va$]
\label{LemVa}
Under the conditions of Theorem~\ref{ThmMain}, we have
\begin{eqnarray}
\prob{[\max_{j \in {\Ssetcomp}}|\Va_j| \geq (1-\delta)\regparn]} &
\rightarrow & 0.
\end{eqnarray}
\end{lem}
\begin{lem}[Control of $\Vb$]
\label{LemVb}
Under the conditions of Theorem~\ref{ThmMain}, we have
\begin{eqnarray}
\mprob[\max_{j \in \Sbar} |\Vb_j| \geq \delta \regparn] & \rightarrow & 0.
\end{eqnarray}
\end{lem}

\begin{lem}[Control of $\Uvar$]
\label{LemUvar}
Under the conditions of Theorem~\ref{ThmMain}, we have
\begin{eqnarray}
\mprob[(\Event(\Uvar))^\mycomplement] \; = \; \mprob[\max_{i \in \Sset}
|\Uvar_i| > \minval] & \rightarrow & 0.
\end{eqnarray}

\end{lem}


\subsection{Proof of Lemma~\ref{LemVa}} 
\label{SecLemVa}

We assume throughout that $\SamCov_{\Sset \Sset}$ is invertible, an
event which occurs with probability $1-o(1)$ under the stated
assumptions (see Lemma~\ref{LemInvertible}).  If we define the
$\numobs$-dimensional vector 
\begin{eqnarray}
\label{EqnDefnKeyvector}
\keyvector & \defn & \Amat_\Sset (\SamCov_{\Sset \Sset})^{-1} \myones,
\end{eqnarray}
then the variable $\Va_j$ can be written compactly as
\begin{equation}
\frac{\Va_j}{\regparn} \; = \; \Amat_j^T \keyvector \, = \,
\sum_{\ell=1}^ \numobs \keyvector_\ell \Amat_{\ell j}.
\end{equation}
Note that each term $\Amat_{\ell j}$ in this sum is distributed as a
mixture variable, taking the value $0$ with probability $1-\spgam$,
and distributed as $N(0, \frac{1}{\spgam})$ variable with probability
$\spgam$.   For each $\ell = 1, \ldots, \numobs$, define the discrete
random variable
\begin{eqnarray}
\label{EqnDefnHvar}
\Hvar_\ell & \edist & \begin{cases} \keyvector_\ell & \mbox{with
probability $\spgam$} \\
0 & \mbox{with probability $1-\spgam$.} 
		      \end{cases}
\end{eqnarray}
For each index $\ell = 1, \ldots, \numobs$, let $\Zvar_{\ell j} \sim
N(0, \frac{1}{\spgam})$.  With these definitions, by construction, we
have
\begin{eqnarray*}
\frac{\Va_j}{\regparn} & \edist & \sum_{\ell=1}^\numobs \Hvar_\ell
\Zvar_{\ell j}.
\end{eqnarray*}
To gain some intuition for the behavior of this sum, note that the
variables $\{ \Zvar_{\ell j}, \ell=1, \ldots, \numobs \}$ are
independent of $\{\Hvar_\ell, \ell=1, \ldots, \numobs \}$.  (In
particular, each $\Hvar_\ell$ is a function of $\Amat_\Sset$, whereas
$\Zvar_{\ell j}$ is a function of $\Amat_{\ell j}$, with $j \notin
\Sset$.)  Consequently, we may condition on $\Hvar$ without affecting
$\Zvar$, and since $Z$ is Gaussian, we have $(\frac{\Va_j}{\regparn}
\, \mid \; \Hvar) \sim N(0, \frac{\|\Hvar\|_2^2}{\spgam})$.
Therefore, if we can obtain good control on the norm $\|\Hvar\|_2$,
then we can use standard Gaussian tail bounds (see
Appendix~\ref{AppTailBounds}) to control the maximum $\max_{j \in
\Sbar} \Va_j/\regparn$.  The following lemma is proved in
Appendix~\ref{AppLemAlpha}:
\begin{lem}
\label{LemAlpha}
Under condition~\eqref{mt00}, then for any fixed $\delta > 0$, we have
\begin{eqnarray*}
\prob \left[\Length{\Hvar}{2}^2 \leq \frac{\gamma \kdim
(1+\delta)}{\numobs} \right] & \geq & 1-
\mathcal{O}(\exp(-\min \{2\log(\pdim - \kdim), \frac{\numobs}{2 \kdim}\}))
\end{eqnarray*} 
\end{lem}

The primary implication of the above bound is that each
$\Va_j/\regparn$ variable is (essentially) no larger than a
$\Normal(0, \frac{\kdim}{\numobs})$ variable. We can then use standard
techniques for bounding the tails of Gaussian variables to obtain good
control over the random variable $\max_{j \in \Sbar}
|\Va_j|/\regparn$.  In particular, by union bound, we have
\begin{eqnarray*}
\mprob[ \max_{j \in \Sbar} |\Va_j| \geq (1-\delta) \regparn] & \leq &
(\pdim - \kdim) \; \mprob[ \sum_{\ell=1}^\numobs \Hvar_{\ell j} Z_j
\geq (1-\delta)]
\end{eqnarray*}
For any $\delta > 0$, define the event $\Tail(\delta) \defn \{
\Length{\Hvar}{2}^2 \leq \frac{\kdim \gamma (1+\delta)}{\numobs}\}$.
Continuing on, we have
\begin{eqnarray*}
\mprob[ \max_{j \in \Sbar} |\Va_j| \geq (1-\delta) \regparn] & \leq &
(\pdim - \kdim) \; \left \{ \mprob[ \sum_{\ell=1}^\numobs \Hvar_{\ell
j} Z_j \geq (1-\delta) \, \mid \, \Tail(\delta)] + \mprob[
(\Tail(\delta)^c)] \right \} \\
& \leq & (\pdim - \kdim) \; \left \{ 2\exp \left(- \frac{\numobs
(1-\delta)^2 }{2\kdim (1+\delta)} \right) +
\mathcal{O}(\exp(-\min{(2\log(\pdim - \kdim), \frac{\numobs}{2
\kdim})})) \right \},
\end{eqnarray*}
where the last line uses a standard Gaussian tail bound (see
Appendix~\ref{AppTailBounds}), and Lemma~\ref{LemAlpha}.  Finally, it
can be verified that under the condition $\numobs > (2+\thmeps) \kdim
\log{(\pdim - \kdim)}$ for some $\thmeps >0$, and with $\delta>0$
chosen sufficiently small, we have $\mprob[ \max_{j \in \Sbar} |\Va_j|
\geq (1-\delta) \regparn] \rightarrow 0$ as claimed.

\subsection{Proof of Lemma~\ref{LemVb}}
\label{SubSectionVjbSmall}

Defining the orthogonal projection matrix $\Projorth{\Sset} \defn
I_{\numobs \times \numobs} - \Amat_\Sset (\Amat^T_\Sset
\Amat_\Sset)^{-1} \Amat^T_\Sset$, we then have
\begin{eqnarray}
\prob[\max_{j \in \Sbar}|\Vb_j| \geq \delta\regparn] & =& 
\prob[\max_{j \in \Sbar} \big| \Amat_j^T
\Projorth{\Sset}(\Wsca/\numobs) \big| \geq \delta \regparn] \nonumber \\
\label{EqnInterTwo}
& \leq & (\pdim - \kdim) \; \prob \left [\big| \Amat_1^T
\Projorth{\Sset}(\Wsca/\numobs) \big| \geq \delta \regparn \right].
\end{eqnarray}

Recall from equation~\eqref{EqnDefnHvar} the representation
$\Amat_{\ell 1} = \Hvar_{ \ell j} Z_{\ell j}$, where $\Hvar_{\ell j}$
is Bernoulli with parameter $\spgam$, and $Z_{\ell j} \sim N(0,
\frac{1}{\spgam})$ is Gaussian.  The variable $\sum_{\ell=1}^\numobs
\Hvar_{\ell j}$ is binomial; define the following event
\begin{eqnarray*}
\Tail & \defn & \left\{ \frac{1}{\numobs} \big| \sum_{\ell=1}^\numobs
\Hvar_{\ell j} - \spgam \numobs \big| \leq \frac{1}{2\sqrt{\kdim}}
\right\}.
\end{eqnarray*}
From the Hoeffding bound (see Lemma~\ref{LemHoeffding}), we have
$\mprob[\Tail^c] \leq 2 \exp(- \frac{\numobs}{2 \kdim})$.  Using this
representation and conditioning on $\Tail$, we have
\begin{eqnarray*}
\prob \left [\big| \Amat_j^T \Projorth{\Sset}(\Wsca/\numobs) \big|
\geq \delta \regparn \right] & \leq & \prob \left [\big|
\frac{1}{\numobs} \sum_{\ell=1}^\numobs \Hvar_{\ell j} Z_{\ell j}
\Projorth{\Sset}(\Wsca)_\ell \big| \geq \delta \regparn \; \mid \;
\Tail \right] + \mprob[\Tail^c] \\
& \leq & \prob \left [\big| \frac{1}{\numobs}
\sum_{\ell=1}^{\numobs(\spgam + \frac{1}{2 \sqrt{\kdim}})} Z_{\ell j}
\Projorth{\Sset}(\Wsca)_\ell \big| \geq \delta \regparn \right] + 2
\exp(- \frac{\numobs}{2 \kdim}),
\end{eqnarray*}
where we have assumed without loss of generality that the first
$\numobs (\spgam + \frac{1}{2 \sqrt{\kdim}})$ elements of $\Hvar$
are non-zero.
Since $\Projorth{\Sset}$ is an orthogonal projection matrix, we
have $\|\Projorth{\Sset}(\Wsca)\|_2 \leq  \|\Wsca\|_2$, so that
\begin{eqnarray}
\label{EqnInterTwoPrime}
\prob \left [\big| \Amat_j^T \Projorth{\Sset}(\Wsca/\numobs) \big|
\geq \delta \regparn \right] & \leq & 
\prob \left [\big|
\frac{1}{\numobs} \sum_{\ell=1}^{\numobs(\spgam + \frac{1}{2
\sqrt{\kdim}})} Z_{\ell j} \Wsca_\ell \big| \geq \delta \regparn
\right] + 2 \exp(- \frac{\numobs}{2 \kdim}),
\end{eqnarray}
Conditioned on $\Wsca$, the random variable $M_j \defn
\frac{1}{\numobs} \sum_{\ell=1}^{\numobs(\spgam + \frac{1}{2
\sqrt{\kdim}})} Z_{\ell j} \Wsca_\ell$ is zero-mean Gaussian with
variance
\begin{equation*}
\nu(\Wsca; \spgam) \defn \frac{1}{\numobs^2 \spgam}  \;
 \sum_{\ell=1}^{\numobs(\spgam + \frac{1}{2 \sqrt{\kdim}})}
 \Wsca_\ell^2.
\end{equation*}
For some $\delta_1 > 0$, define the event
\begin{eqnarray*}
\Tailtwo(\delta_1) & \defn & \left \{ \nu(\Wsca; \spgam) \leq
(1+\delta_1)\frac{\sigma^2}{\numobs \spgam^2} \; (\spgam + \frac{1}{2
\sqrt{\kdim}}) \right \}.
\end{eqnarray*}
Note that $\Exs[\nu(\Wsca; \spgam)] = \frac{\sigma^2}{\numobs \spgam^2}
(\spgam + \frac{1}{2 \sqrt{\kdim}})$.  Since
$\frac{\gamma}{\sigma^2}\sum_{\ell=1}^{\numobs(\spgam + \frac{1}{2 \sqrt{\kdim}})}
\Wsca_\ell^2$ is $\chi^2$ with $d = \numobs(\spgam + \frac{1}{2
\sqrt{\kdim}})$ degrees of freedom, using $\chi^2$-tail bounds (see
Appendix~\ref{AppTailBounds}), we have
\begin{eqnarray*}
\mprob[(\Tailtwo(\delta_1))^c] & \leq &  \exp
\left(-\numobs(\gamma+\frac{1}{2\sqrt{\kdim}}) \frac{3\delta_1^2}{16}
\right).
\end{eqnarray*}
Now, by conditioning on $\Tailtwo(\delta_1)$ and its complement and
using tail bounds on Gaussian variates (see Appendix~\ref{AppTailBounds}),
we obtain
\begin{eqnarray}
\prob \left [\big| \frac{1}{\numobs} \sum_{\ell=1}^{\numobs(\spgam +
\frac{1}{2 \sqrt{\kdim}})} Z_{\ell j} \Wsca_\ell \big| \geq \delta
\regparn \right] & \leq & \prob \left [\big| \frac{1}{\numobs}
\sum_{\ell=1}^{\numobs(\spgam + \frac{1}{2 \sqrt{\kdim}})} Z_{\ell j}
\Wsca_\ell \big| \geq \delta \regparn \, \mid \; \Tailtwo(\delta_1)
\right] + \mprob[(\Tailtwo(\delta_1))^c] \nonumber \\
\label{EqnInterThree}
& \leq & 2 \; \exp \left( - \frac{\numobs \spgam^2 (\delta^2
\regparn^2)}{2 \sigma^2(1+\delta_1) (\spgam + \frac{1}{2 \sqrt{\kdim}})}
\right) + \nonumber \\
&& \hspace{2cm} \exp \left(-\numobs(\gamma+\frac{1}{2\sqrt{\kdim}})
\frac{3\delta_1^2}{16} \right).
\end{eqnarray}


Finally, putting together the pieces from
equations~\eqref{EqnInterThree}, ~\eqref{EqnInterTwoPrime}, and
equation~\eqref{EqnInterTwo}, we obtain that $\prob[\max_{j \in
\Sbar}|\Vb_j| \geq \delta\regparn]$ is upper bounded by
\begin{equation*}
(\pdim - \kdim) \; \left \{ 2 \exp(- \frac{\numobs}{2 \kdim}) + 2 \;
\exp \left( - \frac{\numobs \spgam^2 (\delta^2 \regparn^2)}{2\sigma^2 (1+\delta_1)
(\spgam + \frac{1}{2 \sqrt{\kdim}})} \right) + \exp
\left(-\numobs(\gamma+\frac{1}{2\sqrt{\kdim}}) \frac{3\delta_1^2}{16}
\right) \right \}.
\end{equation*}
The first term goes to zero since $\numobs > (2+\thmeps) \kdim
\log(\pdim - \kdim)$.  The second term goes to zero because eventually $\frac{\spgam^2}{\spgam+\frac{1}{2\sqrt{\kdim}}} > \frac{\gamma}{2}$ (because Condition~\eqref{mt00} implies that $\gamma\sqrt{k} \rightarrow \infty$), and Conditon~\eqref{mt11} implies that $C \numobs \spgam \regparn^2 - \log{(\pdim -\kdim)} \rightarrow \infty$.
Our choice of $\numobs$ and Condition~\eqref{mt00} (which implies that $\gamma \kdim \rightarrow \infty$) is enough for the third term goes to zero.



\subsection{Proof of Lemma~\ref{LemUvar}} 
\label{SubSectionUiSmall}

We first observe that conditioned on $X_{\Sset}$, each $\Uvar_i$ is
Gaussian with mean and variance:
\begin{eqnarray*}
\umean_i & \defn & \Exs [ \Uvar_i \; \mid \; \Amat_{\Sset}] = e_i^T
\big (\frac{1}{\numobs} \Amat_\Sset^T \Amat_\Sset \big)^{-1} \big[-
\regparn \myones \big], \\
\uvariance_i & \defn & \var[\Uvar_i \, \mid \, \Amat_\Sset] =
\frac{\sigma^2}{\gamma\numobs} e_i^T \big(\frac{1}{\numobs} \Amat_\Sset^T
\Amat_\Sset \big)^{-1} e_i
\end{eqnarray*}
Define the upper bounds
\begin{eqnarray*}
\umean^* & \defn & \regparn (1+ \sqrt{\kdim} \; \LHScortwo ) \\
\uvariance^* & \defn & \frac{\sigma^2}{\gamma \numobs} \; \left[1 -\LHScortwo \right]^{-1}
\end{eqnarray*}
and the following event
\begin{eqnarray*}
\Tail(\umean^*, \uvariance^*) & \defn & \{ \max_{i \in \Sset} |
\umean_i| \leq \umean^* \textrm{ and } \max_{i \in \Sset}
|\uvariance_i| \leq \uvariance^* \}.
\end{eqnarray*}
Conditioning on $\Tail$ and its complement, we have
\begin{eqnarray*}
\prob[(\Event(U))^{\mycomplement}] & = & \prob[\frac{1}{\minval}
\max_{i \in \Sset} \Uvar_i| > 1] \\
& \leq & \prob[\frac{1}{\minval}\max_{i \in {\Sset}}|U_i| > 1 \; \mid
\; \Tail(\umean^*, \uvariance^*)] + \prob[(\Tail(\umean^*,
\uvariance^*))^c].
\end{eqnarray*}
Applying Lemma~\ref{LemInvertibleSuper} with $\inverone = 1$ and
$\invertwo = \kdim$, we have $\prob [(\Tail(\umean^*,
\uvariance^*))^c] \leq \kdim \mathcal{O}(k^{-2})$.

We now deal with the first term.  Letting $Y_i \sim \Normal(0,
\uvariance_i)$, and using $\Tail$ as shorthand for the event
$\Tail(\umean^*, \uvariance^*)$, we have
\begin{eqnarray*}
\prob[\frac{1}{\minval}\max_{i \in \Sset} |\Uvar_i| > 1 \; \mid \;
\Tail] & = & \Exs\left\{ \prob \big[\max_{i \in
\Sset} |\Uvar_i| > \minval \; \mid \; X_{\Sset}, \Tail \big] \right \} \\
& \leq & \Exs \left \{ \prob \big[ \max_{i \in \Sset}
\big(|\umean_i|+|Y_i| \big) > \minval \; \mid \; \Amat_{\Sset}, \Tail
\big] \right \} \\
& \leq & \Exs \left \{ \prob \big[\umean^* + \max_{i \in \Sset} |Y_i|
> \minval \; \mid \; \Amat_{\Sset}, \Tail \big] \right \} \\
& = & \Exs \left \{ \prob \big[\frac{1}{\minval}\max_{i \in \Sset}
  |Y_i| > 1 - \frac{\umean^*}{\minval} \; \mid \; \Amat_{\Sset}, \Tail
  \big] \right \}.
\end{eqnarray*}
Condition~\eqref{mt3} implies that $\frac{\umean^*}{\minval}
\rightarrow 0$, so that it suffices to upper bound
\begin{eqnarray*}
\Exs \left \{ \prob \big[\frac{1}{\minval}\max_{i \in \Sset} |Y_i| >
  \frac{1}{2}\; \mid \; \Amat_{\Sset}, \Tail \big] \right \} & \leq &
  \Exs \left \{ \kdim \; \mprob[ |Y^*| \geq \frac{\minval}{2} \;
  \mid \; \Amat_\Sset, \Tail ] \right \} \\
& \leq & 2 \kdim \; \exp \left( - \frac{\minval^2}{8 \uvariance^*} \right).
\end{eqnarray*}
where $Y^* \sim \mathcal{N}(0, \uvariance^*)$, and we have used
standard Gaussian tail bounds (see Appendix~\ref{AppTailBounds}).

It remains to verify that this final term converges to zero.  Taking
logarithms and ignoring constant terms, we have
\begin{eqnarray*}
\log{(\kdim)} (1-\frac{\minval^2}{\log(\kdim) \; 8 \uvariance^*} ) & =
& \log{(\kdim)} \left( 1-\frac{\minval^2 \gamma \numobs \left(1-\LHScortwo
\right)}{8\sigma^2 \log \kdim} \right).
\end{eqnarray*}
We would like to show that this quantity diverges to $-\infty$.
Condition~\eqref{mt00} implies that
\begin{equation*}
{\frac{1}{\gamma} \sqrt{\max\left \{
\frac{\log{(\kdim)}}{\kdim\log{(\pdim - \kdim)}}, \frac{\log
\log(\pdim - \kdim)}{\log(\pdim - \kdim)} \right \}}} \rightarrow 0.
\end{equation*}
Hence, it suffices to show that $\log \kdim \; \left
(1-\frac{\minval^2 \gamma \numobs }{16\sigma^2 \log \kdim} \right)$ diverges
to $-\infty$.
We have

\begin{eqnarray*}
\log (\kdim) \; \left (1-\frac{\minval^2 \gamma \numobs }{16 \log (\kdim)}
\right) & = & \log(\kdim) \; (1-\frac{\minval^2}{\regparn^2}
\frac{\gamma \numobs\regparn^2}{16\sigma^2 \log(\kdim)}) \\
&=& \log(\kdim) \; (1-\frac{\minval^2}{\regparn^2}
\frac{\gamma \numobs\regparn^2}{16\sigma^2 \log(\pdim - \kdim)} \frac{\log(p-k)}{\log{(k)}}) 
\end{eqnarray*}


Condition ($\ref{mt3}$) implies that $\frac{\minval^2}{\regparn^2} \rightarrow \infty$ and 
Condition (\ref{mt11}) states that 
$\frac{\gamma \numobs\regparn^2}{\log(\pdim - \kdim)} \rightarrow \infty$. In our observation model,
$\kdim \leq \frac{\pdim}{2}$, and so the third term is  greater than one.
\\
\\
Therefore, we have that $\prob[\mathcal{E}(U)
^{\mycomplement}]$ tends to zero.

\section{Experimental Results}
\label{SectionExperimental}

In this section, we provide some experimental results to illustrate
the claims of Theorem~\ref{ThmMain}. We consider two different
sparsity regimes, namely linear sparsity ($\kdim=\alpha \pdim$) and
polynomial sparsity ($\kdim=\sqrt{\pdim}$), and we allow $\spgam$ to
converge to zero at some rate.

For all experiments, the additive noise variance is set to
$\sigma^2 = 0.0625$ and we fix the vector $\betastar$ by setting the
first $\kdim$ entries are set to one, and the remaining entries to
zero.  There is no loss of generality in fixing the support in this
way, since the ensemble in invariant under permutations.
 
 Based on
Lemma~\ref{LemUnique}, it suffices to simulate the random variables
$\{\Va_j, \Vb_j, j \in \Sbar \}$ and $\{\Uvar_i, i \in \Sset \}$, and
then check the equivalent conditions~\eqref{nas1} and~\eqref{nas2}.
In all cases, we plot the success probability $\mprob[S(\estim{\beta})
= S(\beta^*)]$ versus the \emph{control parameter} $\contpar(\numobs,
\pdim, \kdim) = \frac{\numobs}{2 \kdim \log(\pdim - \kdim)}$.  Note
that Theorem~\ref{ThmMain} predicts that the Lasso should transition
from failure to success for $\contpar \approx 1$.

\begin{figure}[h]
\begin{center}
\begin{tabular}{ccc}
\widgraph{0.42\textwidth}{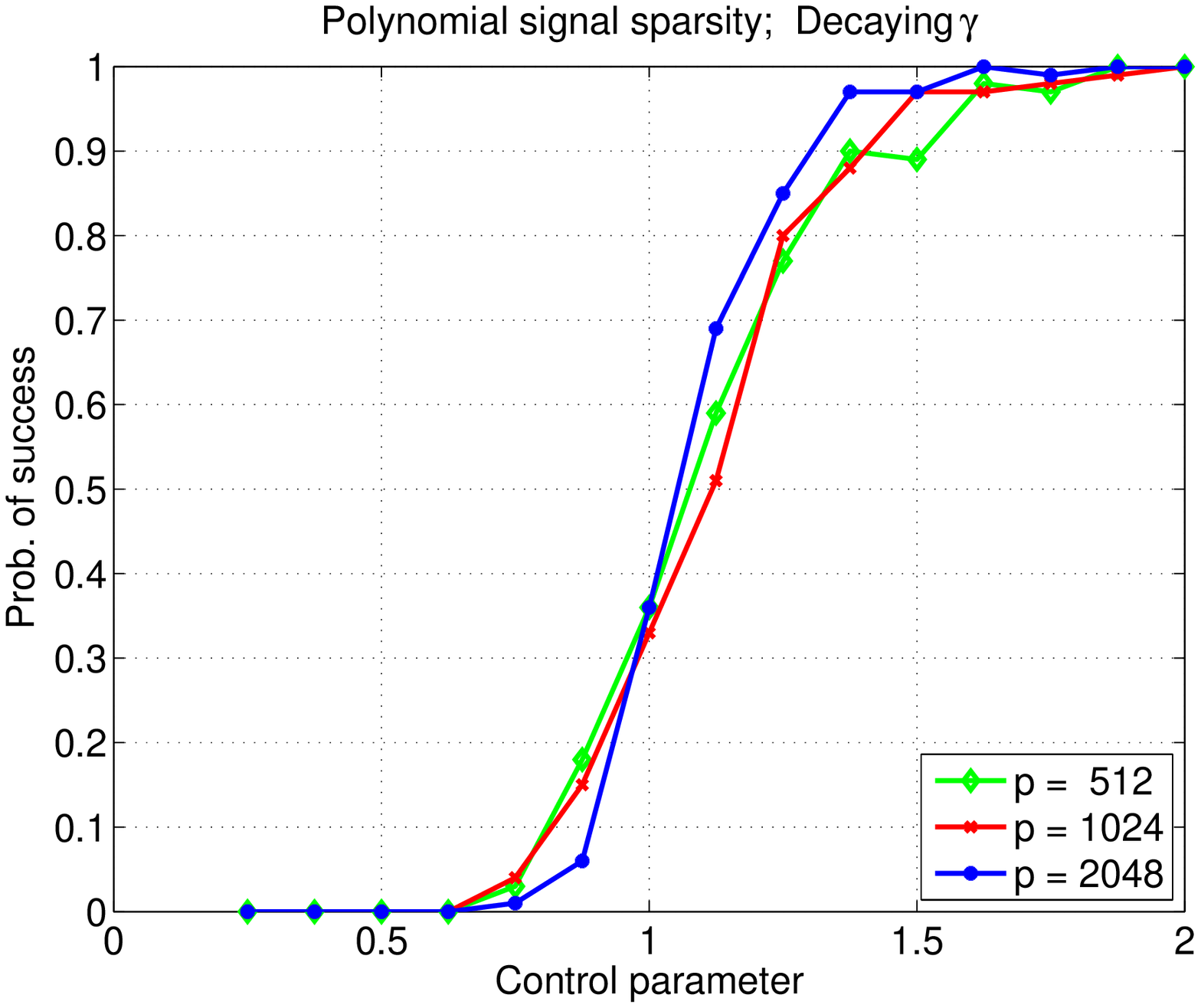} & \hspace*{.2in} &
\widgraph{0.42\textwidth}{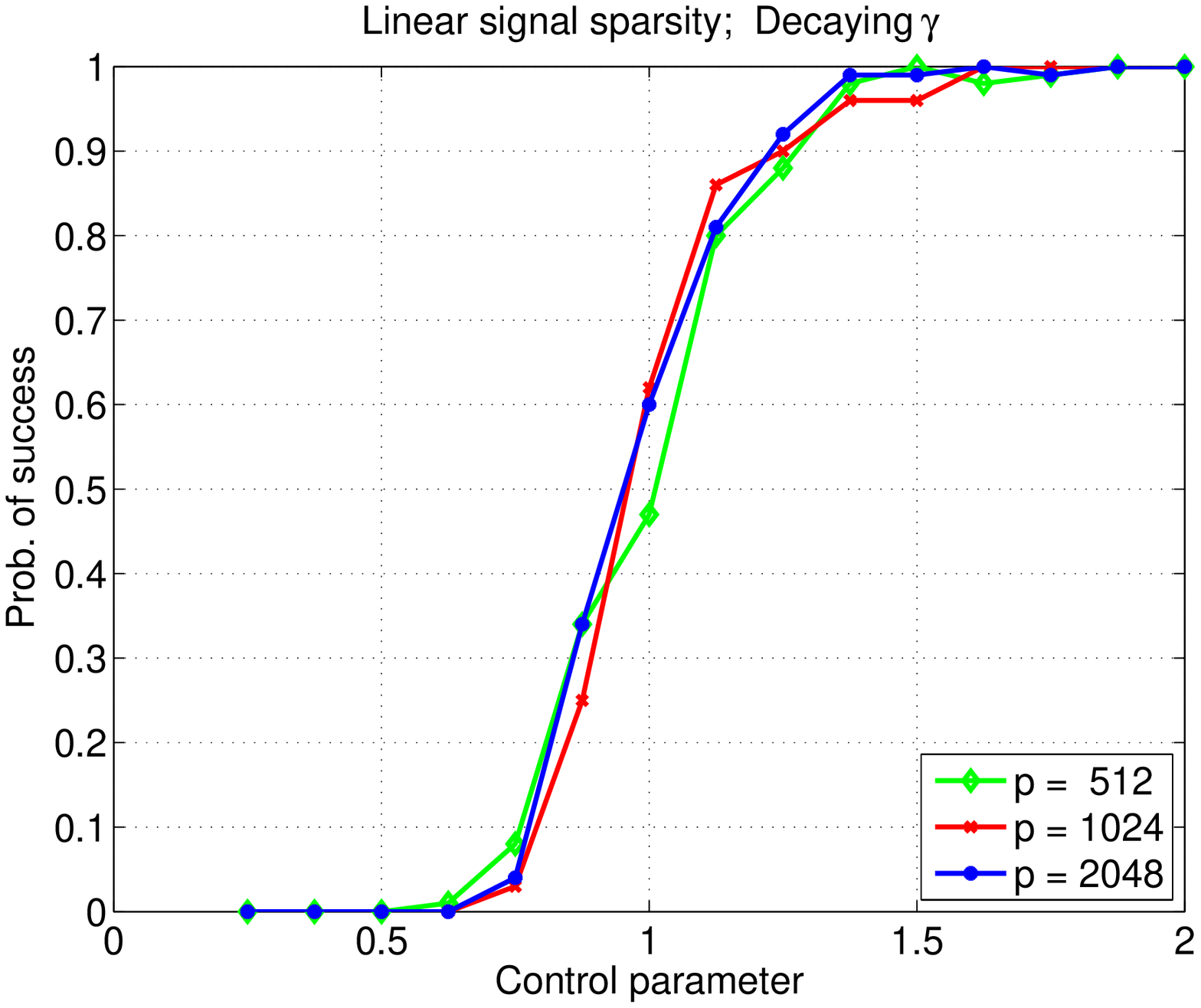} \\
(a) & & (b)
\end{tabular}
\end{center}
\caption{Plots of the success probability $\mprob[\widehat{\Sset} =
\Sset]$ versus the control parameter $\contpar(\numobs, \pdim, \kdim) =
\frac{\numobs}{\kdim \log(\pdim - \kdim)}$ for $\spgam$-sparsified
ensembles, with decaying measurement sparsity $\spgam = \frac{{.5
\log{(p-k)}}}{\sqrt{p-k}}$.  (a) Polynomial signal sparsity $\kdim =
\order(\sqrt{\pdim})$.  (b) Linear signal sparsity $\kdim =
\Theta(\pdim)$.}
\label{FigScaleGamma}
\end{figure}

In Figure~\ref{FigScaleGamma}, the empirical success rate of the
Lasso is plotted against the control parameter $\theta(\numobs, \pdim,
\kdim) = \frac{\numobs}{2 \kdim \log(\pdim - \kdim)}$.  Each panel
shows three curves, corresponding to the problem sizes $\pdim \in
\{512, 1024, 2048\}$, and each point on the curve represents the
average of 100 trials.  For the experiments in
Figure~\ref{FigScaleGamma}, we set $\spgam = 0.5 \frac{\log(\pdim -
\kdim)}{\sqrt{\pdim - \kdim}}$, which converges to zero at a rate
slightly faster than that guaranteed by Theorem~\ref{ThmMain}.
Nonetheless, we still observe the "stacking" behavior around the
predicted threshold $\theta^* = 1$.\\

\section{Discussion}
\label{SectionDiscussion}

In this paper, we have studied the problem of recovery the support set
of a sparse vector $\beta^*$ based on noisy observations.  The main
result is to show that it is possible to ``sparsify'' standard dense
measurement matrices, so that they have a vanishing fraction of
non-zeroes per row, while retaining the same sample complexity (number
of observations $\numobs$) required for exact recovery.  We also
showed that under the support recovery metric and in the presence of
noise, no method can succeed without the number of non-zeroes per
column tending to infinity.  See also the paper~\cite{WanWaiRam08} for
complementary results on the information-theoretic scaling of sparse
measurement ensembles.  

The approach taken in this paper is to find rates which $\gamma$ (as a
function of $\numobs$, $\pdim$, $\kdim$) can safely tend towards zero
while maintaining the same statistical efficiency as dense random
matrices. In various practical settings~\cite{CSCAM}, it may be
preferable to make the measurement ensembles even sparser at the cost
of taking more measurements $\numobs$ and thus decreasing efficiency
relative to dense random matrices. A natural question is the sample
complexity $\numobs(\gamma, \pdim, \kdim)$ in this regime as well.
Finally, this work has focused only on a randomly sparsified matrices,
as opposed to particular sparse designs (e.g., based on LDPC or
expander-type constructions~\cite{Feldman05b,SaBaBa06a, XuHassibi07}).
Although our results imply that exact support recovery with noisy
observations is impossible with bounded degree designs, it would be
interesting to examine the trade-off between other loss functions
(e.g, $\ell_2$ reconstruction error) and sparse measurement designs.

\subsection*{Acknowledgments}  
This work was partially supported by NSF grants CAREER-CCF-0545862 and
DMS-0605165, a Vodafone-US Foundation Fellowship (DO), and a Sloan
Foundation Fellowship (MJW).

\appendix


\section{Standard concentration results}
\label{AppTailBounds}

In this appendix, we collect some tail bounds used repeatedly
throughout this paper.

\begin{lem}[Hoeffding bound~\cite{Hoeffding63}]
\label{LemHoeffding}
Given a binomial variate $Z \sim \Bin(n, \gamma)$, we have for any
$\delta > 0$
\begin{eqnarray*}
\prob{[| Z -\gamma \numobs | \geq \delta \numobs]} & \leq & 2 \, \exp
\big(-2\numobs \delta^2 \big).
\end{eqnarray*}
\end{lem}

\begin{lem}[$\chi^2$-concentration~\cite{Johnstone01}]
\label{LemChiConcentrate}
Let $X \sim \chi^2_m$ be a chi-squared variate with $m$ degrees
of freedom.  Then for all $\frac{1}{2} > \delta \geq 0$, we have
\begin{eqnarray*}
\prob{[ X-m \geq \delta m]} & \leq & \exp \left(-
\frac{3}{16} m \delta^2 \right).
\end{eqnarray*}
\end{lem}

We will also find the following standard Gaussian tail
bound~\cite{LedTal91} useful:
\begin{lem}[Gaussian tail behavior]
\label{Lem1}
Let $V \sim \Normal(0, \sigma^2)$ be a zero-mean Gaussian with
variance $\sigma^2$.  Then for all $\delta > 0$, we have 
\begin{eqnarray*}
\mprob[ |V| > \delta] & \leq & 2 \; \exp
\big(-\frac{\delta^2}{2\sigma^2} \big).
\end{eqnarray*}

\end{lem}

\section{Convex optimality conditions}

\subsection{Proof of Lemma~\ref{LemUnique}}
\label{AppLemUnique}

Let $f(\beta) \defn \frac{1}{2\numobs} \| \Ysca - \Amat \beta \|_2^2 +
\regparn \|\beta\|_1$ denote the objective function of the
Lasso~\eqref{EqnDefnLasso}. By standard convex optimality
conditions~\cite{Rockafellar}, a vector $\betahat \in \real^\pdim$ is
a solution to the Lasso if and only if $0 \in \mathbb{R}^{\pdim}$ is
an element of the subdifferential of $f(\beta)$ at $\betahat$.  These
conditions lead to
\begin{eqnarray*}
\frac{1}{\numobs}\Amat^T (\Amat \betahat - \Ysca) +\regparn \dualvec &
= & 0,
\end{eqnarray*}
where the dual vector $\dualvec \in \real^\pdim$ is an element of the
subdifferential of the $\ell_1$-norm, given by
\begin{eqnarray*}
\partial \|\betahat\|_1 &=& \left\{ z \in \real^{\pdim} \; \mid \; z_i
  = \sign(\betahat_i) \textrm{ if } \betahat_i \neq 0, \qquad z_i \in
  [-1, 1] \textrm{ otherwise} \right\}.
\end{eqnarray*}

Now suppose that we are given a pair $(\estim{\beta}, \dualvec) \in
\real^\pdim \times \real^\pdim$ that satisfy the assumptions of
Lemma~\ref{LemUnique}.  Condition~\eqref{EqnZeroSub} is equivalent to
$(\estim{\beta}, \dualvec)$ satisfying the zero subgradient condition.
Conditions~\eqref{EqnUniqueA},~\eqref{EqnUniqueC}
and~\eqref{EqnUniqueD} ensure that $\dualvec$ is an element of the
subdifferential of the $\ell_1$-norm at $\estim{\beta}$.  Finally,
conditions~\eqref{EqnUniqueB} and~\eqref{EqnUniqueD} ensure that
$\estim{\beta}$ correctly specifies the signed support.

It remains to verify that $\estim{\beta}$ is the \emph{unique} optimal
solution.  By Lagrangian duality, the Lasso
problem~\eqref{EqnDefnLasso} (given in penalized form) can be written
as an equivalent constrained optimization problem over the ball
$\|\beta\|_1 \leq C(\regpar_\numobs)$, for some constant
$C(\regpar_\numobs) < +\infty$.  Equivalently, we can express this
single $\ell_1$-constraint as a set of $2^\pdim$ linear constraints
$\signvec^T \beta \leq C$, one for each sign vector $\signvec \in
\{-1, +1\}^\pdim$. The vector $\dualvec$ can be written as a convex
combination $\dualvec = \sum_{\signvec} \alpha^*_\signvec \signvec$,
where the weights $\alpha^*_\signvec$ are non-negative and sum to one.
By construction of $\estim{\beta}$ and $\dualvec$, the weights
$\alpha^*$ form an optimal Lagrange multiplier vector for the problem.
Consequently, any other optimal solution---say $\wtil{\beta}$---must
also minimize the associated Lagrangian
\begin{eqnarray*}
L(\beta; \alpha^*) & = & f(\beta) + \sum_{\signvec} \alpha^*_\signvec
\left [\signvec^T\beta - C \right],
\end{eqnarray*}
and satisfy the complementary slackness conditions $\alpha^*_\signvec
\left(\signvec^T \wtil{\beta} -C \right)= 0$.  Note that these
complementary slackness conditions imply that $\dualvec^T \wtil{\beta}
= C$.  But this can only happen if $\wtil{\beta}_j = 0$ for all
indices where $|\dualvec_j| < 1$.  Therefore, any optimal solution
$\wtil{\beta}$ satisfies $\wtil{\beta}_{\Sbar} = 0$.  Finally, given
that all optimal solutions satisfy $\beta_{\Sbar} = 0$, we may
consider the restricted optimization problem subject to this set of
constraints.  If the Hessian submatrix $\SamCov_{\Sset \Sset}$ is
strictly positive definite, then this sub-problem is strictly convex,
so that $\estim{\beta}$ must be the unique optimal solution, as
claimed.

\subsection{Derivation of $\{\Va_j, \Vb_j, \Uvar_i \}$}
\label{SecVUvar}

In this appendix, we derive the form of the $\{\Va_j, \Vb_j \}$ and
$\{\Uvar_i \}$ variables defined in equations~\eqref{EqnDefnVa}
through~\eqref{EqnDefnUvar}.  We begin by writing the zero
sub-gradient condition in a block-form, and substituting the relations
specified in conditions~\eqref{EqnUniqueA} and~\eqref{EqnUniqueB}:
\begin{eqnarray*}
\begin{bmatrix} \SamCov_{\Sset \Sset} & \SamCov_{\Sset \Sbar} \\
\SamCov_{\Sbar \Sset} & \SamCov_{\Sbar \Sbar} 
\end{bmatrix}
\begin{bmatrix}
\estim{\beta}_\Sset - \beta^*_\Sset \\ 0
\end{bmatrix}
+ \begin{bmatrix}  \frac{1}{\numobs} \Amat_\Sset^T W \\
\frac{1}{\numobs} \Amat_{\Sbar}^T W \end{bmatrix}
+
\regparn
\begin{bmatrix}
\sign(\beta^*_\Sset) \\ \dualvec_{\Sbar}
\end{bmatrix}  & = & 0.
\end{eqnarray*}
By solving the top block, we obtain
\begin{eqnarray*}
\Uvar \defn \estim{\beta}_\Sset - \beta^*_\Sset & = & -\inv{\SamCov_{\Sset \Sset}}
\left \{ \frac{1}{\numobs} \Amat^T_\Sset \Wsca + \regparn
\sign(\beta^*_\Sset) \right \}.
\end{eqnarray*}
By back-substituting this relation into the lower block, we can solve
explicitly for $\dualvec_{\Sbar}$; doing so yields that
$\dualvec_{\Sbar} = \Va + \Vb$, where the $(\pdim -\kdim)$-vectors are
defined in equations~\eqref{EqnDefnVa} and~\eqref{EqnDefnVb}.


%


\section{Proof of Lemma~\ref{LemAlpha}}
\label{AppLemAlpha}

Let $Z \in \real^{\numobs \times \numobs}$ denote a $\numobs \times
\numobs$ matrix, for which the off-diagonal elements $Z_{ij} = 0$ for
all $i \neq j$, and the diagonal elements $Z_{ii} \sim \Ber(\spgam)$
are i.i.d.  With this notation, we can write $\Hvar \edist Z
\keyvector$.  Using the definition~\eqref{EqnDefnKeyvector} of
$\keyvector$, we have
\begin{eqnarray*}
\Length{\Hvar}{2}^2 &=&  \|Z \keyvector\|_2^2 \\
& = & \|Z \; \frac{\Amat_\Sset}{\numobs} (\SamCov_{\Sset \Sset})^{-1}
\myones \|_2^2 \\
& = & \myones^T (\SamCov_{\Sset \Sset})^{-1} (Z
\frac{\Amat_\Sset}{\numobs})^T (Z \frac{\Amat_\Sset}{\numobs})
(\SamCov_{\Sset \Sset})^{-1} \myones \\
 &=& \frac{\spgam}{\numobs} \myones^T (\SamCov_{\Sset \Sset})^{-1}
\underbrace{\left \{ \frac{1}{\spgam \numobs} \sum_{i=1}^\numobs
\Ind[Z_{ii} =1] \; x_i x_i^T \right \}} (\SamCov_{\Sset \Sset})^{-1}
\myones \\
& & \qquad \qquad \qquad \qquad  \Randmat(Z)
\end{eqnarray*}
where $x_i$ is the $i^{th}$ row of the matrix $\Amat_\Sset$.  From
Lemma~\ref{LemInvertibleSuper} with $\invertwo = 1$ and
$\inverone = (\pdim - \kdim)$,   we have
\begin{eqnarray}
\label{EqnBoundOne}
\mprob \left[ \Spectral{\myinv{\SamCov_{\Sset \Sset}}} \geq
\fone(\pdim, \kdim, \spgam) \right] & \leq & \frac{1}{(\pdim -
\kdim)^2}
\end{eqnarray}
where $\fone(\pdim,\kdim,\gamma) \defn \ellone$.

Next we control the spectral norm of the random matrix $\Randmat(Z)$,
conditioned on the total number $\sum_{i=1}^\numobs Z_{ii}$ of
non-zero entries.  In particular, applying
Lemma~\ref{LemInvertibleSuper} with $\inverone = \pdim - \kdim$, and
$\invertwo = 1$, we have
\begin{eqnarray}
\label{EqnInterBound}
\prob \left[\Length{\Randmat{Z}}{2} \geq \frac{z}{\numobs\gamma}
\big[1+\LHS{\frac{z}{\numobs}} \big] \; \mid \; \sum_{i=1}^\numobs
Z_{ii} = z \right] & \leq & \frac{1}{(p-k)^2},
\end{eqnarray}
as long as $\kdim\frac{z}{\numobs} \rightarrow \infty$.  

The next step is to deal with the conditioning.  Define the event
\begin{eqnarray*}
\Tail(\kdim, \spgam) & \defn & \left \{ Z \; \mid \; \spgam -
\frac{1}{\sqrt{2\kdim}} \leq \frac{1}{\numobs} \sum_{i=1}^\numobs
Z_{ii} \leq \spgam + \frac{1}{2\sqrt{\kdim}} \right \}.
\end{eqnarray*}
Defining the function
\begin{eqnarray*}
\ftwo(\pdim, \kdim, \gamma) & \defn & \elltwo,
\end{eqnarray*}
we have
\begin{eqnarray}
\mprob[\Spectral{\Randmat(Z)} \geq \ftwo(\pdim, \kdim, \spgam)] & \leq
& \mprob[\Spectral{\Randmat(Z)} \geq \ftwo(\pdim, \kdim, \spgam) \;
\mid \; \Tail(\kdim, \spgam)] + \mprob[(\Tail(\kdim, \spgam))^c] \nonumber \\
& \leq & \vershprobb{2} + 2 \exp(-\frac{\numobs}{2\kdim}) \nonumber \\
\label{EqnBoundTwo}
& \leq & 3 \exp( - \min \{ 2 \log(\pdim-\kdim), \frac{\numobs}{2
\kdim} \}),
\end{eqnarray}
where we have used the bound~\eqref{EqnInterBound}, and the Hoeffding
bound (see Lemma~\ref{LemHoeffding}).

Combining the bounds~\eqref{EqnBoundOne} and~\eqref{EqnBoundTwo}, we
conclude that as long as $\gamma \kdim \rightarrow \infty$, then:
\begin{eqnarray*}
\mprob\left[ \Spectral{\inv{\SamCov} \Randmat(Z) \inv{\SamCov}} \geq
\fone^2 \ftwo \right] & \leq & 4 \exp( - \min \{ 2 \log(\pdim-\kdim),
\frac{\numobs}{2\kdim} \}).
\end{eqnarray*}
Since $\|\myones\|_2 = \sqrt{\kdim}$, we have
\begin{eqnarray*}
\mprob[ \|\Hvar\|_2^2 \geq \frac{\spgam \kdim}{\numobs} \fone^2 \ftwo]
& \leq & 4 \exp( - \min \{ 2 \log(\pdim-\kdim), \frac{\numobs}{2\kdim} \}).
\end{eqnarray*}
To conclude the proof, we note that assumption~\eqref{mt00} implies
that both $\fone(\pdim, \kdim, \gamma)$ and $\ftwo(\pdim, \kdim,
\gamma)$ converge to $1$ as $(\pdim, \kdim, \spgam)$ scale.
In particular, for any fixed $\delta > 0$, we have $\fone^2 \ftwo < (1+\delta)$
for $(\pdim, \kdim)$ sufficiently large, so that Lemma~\ref{LemAlpha}
follows.


\section{Singular values of sparsified matrices}
\label{AppLemInvertible}

Let $\invertwo(p, k) \in (0,1]$ and $\inverone(p, k) \in \{1, 2, 3, \ldots\}$ be functions.  Let
$\Amat$ be an $\invertwo \numobs \times \kdim$ random matrix with
i.i.d. entries $\Amat_{ij}$ distributed according to the
$\spgam$-sparsified ensemble~\eqref{EqnSparseMeas}.
\begin{lem}
\label{LemInvertibleSuper}
Suppose that $\numobs \geq (2+\nu) \kdim \log{(\pdim - \kdim)}$ for
some $\nu > 0$.  If as $\kdim, \pdim-\kdim, \rightarrow \infty$
\begin{eqnarray*}
T(\gamma, \kdim, \pdim, \invertwo, \inverone) & \defn &
 \frac{1}{\gamma} \sqrt{\max \left \{
 \frac{\log{(\inverone)}}{\invertwo \kdim \log{(\pdim - \kdim)}},
 \frac{\log [\invertwo \log(\pdim - \kdim)]}{\invertwo \log(\pdim -
 \kdim)} \right \}} \; \longrightarrow \; 0
\end{eqnarray*}
then for some constant $C \in (0, \infty)$, we have
\begin{eqnarray}
\label{EqnSuper}
\mprob \left[ \sup_{\|u\|_2 = 1} \big | \frac{1}{\sqrt{\invertwo
\numobs}} \|\Amat u\|_2 -1 \big | \geq C \, T(\gamma, \kdim, \pdim,
\invertwo, \inverone) \right] & \leq & \mathcal{O}(\frac{1}{\inverone^2}),
\end{eqnarray}
\end{lem}

Note that Lemma~\ref{LemInvertibleSuper} with $\invertwo = 1$ and
$\inverone = \pdim - \kdim$ implies that $\SamCov = \frac{1}{\numobs}
\Amat_\Sset^T \Amat_\Sset$ is invertible with probability greater than
$1- \mathcal{O}(\frac{1}{(\pdim-\kdim)^2})$, there establishing Lemma~\ref{LemInvertible}.
Other settings in which this lemma is applied are $(\invertwo, \inverone) = (\spgam, \pdim - \kdim)$
and $(\invertwo, \inverone) =
(1, \kdim)$.  The remainder of this section is devoted to the proof
of Lemma~\ref{LemInvertibleSuper}.

\subsection{Bounds on expected values}

Let $\Amat \in \real^{{\invertwo \numobs} \times \kdim}$ be a random
matrix with i.i.d. entries, of the sparsified Gaussian form
\begin{eqnarray*}
\Amat_{ij} & \sim & (1-\spgam) \delta_X(0) + \spgam N(0, \frac{1}{\spgam}).
\end{eqnarray*}
Note that $\Exs[X_{ij}] = 0$ and $\var(X_{ij}) = 1$ by construction.

We follow the proof technique outlined in~\cite{VershNew}. We first
note the tail bound:
\begin{lem}
\label{LemSimpTail}
Let $Y_1, \ldots, Y_d$ be i.i.d. samples of the $\spgam$-sparsified
ensemble.  Given any vector $a \in \real^d$ and $t > 0$, we have
$\mprob[\sum_{i=1}^d a_i Y_i > t] \leq \exp \left(- \frac{\spgam
t^2}{2 \|a\|^2_2} \right)$.
\end{lem}
To establish this bound, note that each $Y_i$ is dominated
(stochastically) by the random variable $Z \sim N(0,
\frac{1}{\spgam})$.  In particular, we have
\begin{eqnarray*}
\MGF_{Y_i}(\lambda) & = & \Exs[\exp(\lambda Y_i)] \; = \; (1-\spgam) +
\spgam \Exs[\exp(\lambda Z)] \leq \exp(\lambda^2/2 \spgam).
\end{eqnarray*}

Now let us bound the maximum singular value $\sval_\kdim(\Amat)$ of
the random matrix $\Amat$.  Letting $S^{d-1}$ denote the $\ell_2$ unit
ball in $d$ dimensions, we begin with the variational representation
\begin{eqnarray*}
\sval_k(\Amat) & = & \max_{u \in S^{\kdim-1}} \| \Amat u \| \\
& = & \max_{v \in S^{{\invertwo \numobs} -1}} \; \max_{u \in
S^{\kdim-1}} v^T \Amat u.
\end{eqnarray*}
For an arbitrary $\epsilon \in (0, 1)$, we can find $\epsilon$-covers
(in $\ell_2$ norm) of $S^{{\invertwo \numobs} -1}$ and $S^{\kdim-1}$
with \mbox{$M_{\invertwo \numobs}(\epsilon) = (3/\epsilon)^{{\invertwo
\numobs}}$} and $M_\kdim(\epsilon) = (3/\epsilon)^\kdim$ points
respectively~\cite{Matousek}.  Denote these covers by $C_{\invertwo
\numobs}(\epsilon)$ and $C_\kdim(\epsilon)$ respectively.  A standard
argument shows that for all $\epsilon \in (0,1)$, we have
\begin{eqnarray*}
\|\Amat \|_2 & \leq & \frac{1}{(1-\epsilon)^2} \max_{u_\alpha \in
C_\kdim(\epsilon)} \; \max_{v_\beta \in C_{\invertwo
\numobs}(\epsilon)} v_\beta^T \Amat u_\alpha.
\end{eqnarray*}
Let us analyze the maximum on the RHS: for a fixed pair $(u,v)$ in our
covers, we have
\begin{eqnarray*}
u^T \Amat v & = & \sum_{i=1}^{\invertwo \numobs} \sum_{j=1}^\kdim
X_{ij} u_i v_j.
\end{eqnarray*}
Let us apply Lemma~\ref{LemSimpTail} with $d = {\invertwo \numobs} \kdim$, and
weights $a_{ij} = u_i v_j$.  Note that we have
\begin{eqnarray*}
\|a \|^2_2 = & = & \sum_{i,j} a_{ij}^2 \; = \; \sum_i u_i^2 (\sum_j
v_j^2) \; = \; 1
\end{eqnarray*}
since each $u$ and $v$ are unit norm.  Consequently, for any fixed
$u, v$ in the covers, we have
\begin{eqnarray*}
\mprob[u^T \Amat v > t] & \leq & \exp \left( - \frac{\spgam t^2}{2}
\right)
\end{eqnarray*}
By the union bound, we have
\begin{eqnarray*}
\mprob\left[\max_{u_\alpha \in C_\kdim(\epsilon)} \; \max_{v_\beta \in
C_{\invertwo \numobs}(\epsilon)} v_\beta^T \Amat u_\alpha > t \right] & \leq &
M_\kdim(\epsilon) M_{\invertwo \numobs}(\epsilon) \exp \left(- \frac{\spgam
t^2}{2} \right) \\
& \leq & \exp \left((\kdim + {\invertwo \numobs}) \log(3/\epsilon) -
\frac{\spgam t^2}{2} \right).
\end{eqnarray*}
By choosing $\epsilon = \frac{1}{2}$ and $t = \sqrt{ \frac{4}{\spgam}
(\kdim + {\invertwo \numobs}) \log 6}$, we can conclude that
\begin{eqnarray*}
\sval_1(\Amat)/\sqrt{{\invertwo \numobs}} \; = \;
\|\Amat\|_2/\sqrt{{\invertwo \numobs}} & \leq & C
\sqrt{\frac{1}{\spgam}} \; \sqrt{1 + \frac{\kdim}{{\invertwo
\numobs}}}
\end{eqnarray*}
w.p. $1 - \exp(-(\kdim+{\invertwo \numobs}) \log 6)$.    Note
that
\begin{eqnarray*}
\frac{\kdim}{\invertwo \numobs} & = & \mathcal{O} \left( \frac{1}{(2 + \nu)
\invertwo \log(\pdim -\kdim)} \right) \rightarrow 0,
\end{eqnarray*}
since $\frac{\log [\invertwo \log(\pdim-\kdim)]}{\invertwo \log{(p-k)}}
\rightarrow 0$, which implies that $\invertwo \log{(\pdim - \kdim)}
\rightarrow \infty$.

Consequently, we can conclude that
\begin{eqnarray*}
\|\Amat\|_2/\sqrt{{\invertwo \numobs}} & \leq &
\mathcal{O}(1/\sqrt{\spgam})
\end{eqnarray*}
w.p. one as ${\invertwo \numobs}, \kdim \rightarrow \infty$.  Although
this bound is essentially correct for a $\mathcal{N}(0,
\frac{1}{\spgam})$ ensemble with $\spgam$ \emph{fixed}, it is very
crude for the sparsified case with $\spgam \rightarrow 0$, but will
useful in obtaining tighter control on $\sval_1(\Amat)$ and
$\sval_k(\Amat)$ in the sequel.

\subsection{Tightening the bound}

For a given $u \in S^{\kdim -1}$, consider the random variable
$\|\Amat u\|_2^2 \defn \sum_{i=1}^{\invertwo \numobs} (\Amat u)^2_i$.
We first claim that each variate $Z_i = (\Amat u)_i^2$ is
subexponential:
\begin{lem}
For any $t > 0$, we have $\mprob[Z_i > t] \leq 2 \exp
\left(-\frac{\spgam t}{2} \right)$.
\begin{proof}
We can write $(\Amat u)_i = \sum_{j=1}^\kdim \Amat_{ij} u_j$ where
$\|u\|_2=1$.  Hence, from Lemma~\ref{LemSimpTail}, we have
\begin{eqnarray*}
\mprob[\sum_{j=1}^\kdim \Amat_{ij} u_j > \delta] & \leq &
\exp(-\frac{\spgam \delta^2}{2}).
\end{eqnarray*}
By symmetry, we have $\mprob[Z_i > t] = \mprob[|\sum_{j=1}^\kdim
\Amat_{ij} u_j| > \sqrt{t}] \; \leq \; 2 \exp(-\frac{\spgam t}{2})$ as
claimed.
\end{proof}
\end{lem}

Now consider the event
\begin{eqnarray*}
\mprob \left[ \left| \frac{\|\Amat u\|^2_2}{{\invertwo \numobs}} -1
\right| > \delta \right] & = & \mprob \left[
\left|\sum_{i=1}^{\invertwo \numobs} Z_i - \Exs[\sum_{i=1}^{\invertwo
\numobs} Z_i] \right| > \delta {\invertwo \numobs} \right]
\end{eqnarray*}
We may apply Theorem 1.4 of Vershynin~\cite{VershNew} with $b = 8
{\invertwo \numobs}/\spgam^2$ and $d = 2/\spgam$.  Hence, we have
$4b/d = 16 {\invertwo \numobs}/\spgam$, which grows at least linearly
in ${\invertwo \numobs}$.  Hence, for any $\delta > 0$ less than $16
{\invertwo \numobs}/\spgam$ (we will in fact take $\delta \rightarrow
0$), we have
\begin{eqnarray*}
\mprob \left[ \left| \frac{\|\Amat u\|^2_2}{{\invertwo \numobs}} -1
\right| > \delta \right] & \leq & 2 \exp \left( -\frac{\delta^2
{(\invertwo \numobs)}^2}{256 {\invertwo \numobs}/\spgam^2} \right) \;
= \; 2 \exp \left( -\frac{\spgam^2 \; \delta^2 {\invertwo \numobs}
}{256} \right).
\end{eqnarray*}
Now take an $\epsilon$-cover of the $\kdim$-dimensional $\ell_2$ ball,
say with $N(\epsilon) \; = \; (3/\epsilon)^\kdim$ elements.  By union
bound, we have
\begin{eqnarray*}
\mprob \left [\inf_{i=1, \ldots, N(\epsilon)} \frac{\|\Amat
u_i\|^2_2}{{\invertwo \numobs}} < 1 - \delta \right] & \leq & \exp
\left(-\frac{\spgam^2 \; \delta^2 {\invertwo \numobs} }{256} + \kdim
\log (3/\epsilon) \right)
\end{eqnarray*}
Now set
\begin{eqnarray*}
\delta = \frac{\sqrt{2}}{\spgam} \sqrt{ \frac{256 f(\kdim, \pdim) \kdim
\log(3/\epsilon)}{{\invertwo \numobs}}},
\end{eqnarray*}
where $f(\kdim, \pdim) \geq 1$ is a function to be specified.  Doing
so yields that the infimum is bounded by $1 + \delta$ with probability
$1 - \exp(-\kdim f(\kdim, \pdim) \log(3/\epsilon))$.  (Note that the
choice of $f(\kdim, \pdim)$ influences the rate of convergence, hence
its utility.)

For any element $u \in S^{\kdim -1}$, we have some $u_i$ in the cover,
and moreover
\begin{eqnarray*}
\left| \| \Amat u\|^2 - \|\Amat u_i\|^2 \right| & = & \left|
\left\{\|\Amat u\| - \|\Amat u_i\| \right\} \; \left\{\|\Amat u\| +
\|\Amat u_i\| \right\} \right| \\
& \leq & \;  \left|
\left\{\|\Amat u\| - \|\Amat u_i\| \right\} \right|          \; ( 2 \|\Amat\|) \;  \; \\
& \leq & (\| \Amat\| \; \|u - u_i\| ) \; \; ( 2 \|\Amat\|) \; \leq \;
2 \|\Amat\|^2 \epsilon
\end{eqnarray*}
From our earlier result, we know that $\|\Amat\|^2 =
\mathcal{O}({\invertwo \numobs}/\spgam)$ with probability
$1-\exp(\log{6} (\kdim+{\invertwo \numobs}))$.  Putting together the
pieces, we have that the bound
\begin{eqnarray*}
\frac{1}{{\invertwo \numobs}} \inf_{u \in S^{k-1}} \|\Amat u\|^2 &
\geq & 1 + \delta + C_2 \epsilon/\spgam \; = \; 1 + \frac{2}{\spgam}
\sqrt{ \frac{32 f(\kdim, \pdim) \kdim \log(3/\epsilon)}{{\invertwo
\numobs}}} + \frac{C_2}{\spgam} \epsilon,
\end{eqnarray*}
for some constant $C_2 > 0$ independent of ${\invertwo \numobs},
\kdim, \spgam$, holds with probability at least
\begin{equation}
\label{EqnProbRates}
\min\{1 - \exp(-\kdim f(\kdim, \pdim) \log(3/\epsilon)),
1-\exp(-\log{6}(\kdim+{\invertwo \numobs})) \},
\end{equation}
Now set $\epsilon = 3 \kdim/{\invertwo \numobs}$, so that we have
w.h.p.
\begin{eqnarray*}
\label{EqnKeyBound}
\frac{1}{{\invertwo \numobs}} \inf_{u \in S^{k-1}} \|\Amat u\|^2 &
\geq & 1 - \frac{C_3}{\spgam} \sqrt{ f(\kdim, \pdim)
\frac{\kdim}{{\invertwo \numobs}} \log( \frac{{\invertwo
\numobs}}{\kdim})}
\end{eqnarray*}
(Note that we have utilized the fact that both $ \sqrt{ f(\kdim,
\pdim) \frac{\kdim}{{\invertwo \numobs}} \log( \frac{{\invertwo
\numobs}}{\kdim})}$ and $\frac{\kdim}{{\invertwo \numobs}} \rightarrow
0$, but the former more slowly than the latter.)

Since $\kdim/{\invertwo \numobs} \rightarrow 0$, this quantity will go
to zero, as long as $f(\kdim, \pdim)$ remains fixed, or scales slowly
enough.  To understand how to choose $f(\kdim, \pdim)$, let us
consider the rate of convergence~\eqref{EqnProbRates}.  To establish
the claim~\eqref{EqnSuper}, we need rates fast enough to dominate a
$\log(\inverone)$ term in the exponent, which guides our choice of
$f(\kdim, \pdim)$.  Recall that we are seeking to prove a scaling of
the form ${\numobs} = \Theta(\kdim \log(\pdim - \kdim))$, so that our
requirement (with $\epsilon = 3 \kdim/{\invertwo \numobs} =
\frac{3}{\invertwo \log(\pdim - \kdim)}$) is equivalent to the
quantity
\begin{eqnarray*}
\kdim f(\kdim, \pdim) \log(3/\epsilon) - \log(\inverone) & = & \kdim
f(\kdim, \pdim) \log [\invertwo \log(\pdim - \kdim)] - \log(\inverone)
\end{eqnarray*}
tending to infinity.  First, if $\kdim > \frac{\log(\inverone)} {\log
[\invertwo \log(\pdim - \kdim)]}$, then we may simply set $f(\kdim,
\pdim) = 2$.  Otherwise, if $\kdim \leq \frac{ \log(\inverone)}{\log
\invertwo \log(\pdim - \kdim)}$, then we may set
\begin{eqnarray*}
f(\kdim, \pdim) = 2\frac{\log(\inverone)}{\kdim \log \invertwo
\log(\pdim - \kdim)} \geq 1.
\end{eqnarray*}
If $f(\kdim, \pdim) = 2$, then we have
\begin{eqnarray*}
f(\kdim, \pdim) \frac{\kdim}{{\invertwo \numobs}} \log(
\frac{{\invertwo \numobs}}{\kdim}) & = & 2 \frac{ \log [\invertwo
\log(\pdim - \kdim)]}{\invertwo \log(\pdim - \kdim)} \; \rightarrow \;
0.
\end{eqnarray*}
In the other case, if $\kdim \leq \frac{ \log(\inverone)}{\log
\invertwo \log(\pdim - \kdim)}$, we have
\begin{eqnarray*}
f(\kdim, \pdim) \frac{\kdim}{{\invertwo \numobs}} \log(
\frac{{\invertwo \numobs}}{\kdim}) & \leq & 2
\frac{\log(\inverone)}{\kdim \log \invertwo \log(\pdim - \kdim)}
\frac{1}{\invertwo \log(\pdim - \kdim)} \log \invertwo \log (\pdim -
\kdim) \; = \; \frac{2}{\kdim}\frac{\log{\inverone}}{\invertwo
\log{(\pdim - \kdim)}} \rightarrow 0,
\end{eqnarray*}
which again follows from the assumptions in Lemma~\ref{LemInvertibleSuper}.

Recalling the definition of $T(\gamma, \kdim, \pdim, \invertwo, \inverone)$ from Lemma~\ref{LemInvertibleSuper}, we can summarize both cases can be summarized cleanly by saying that with probability
greater than $1- \frac{1}{\inverone^2}$:
\begin{eqnarray*}
\frac{1}{{\invertwo \numobs}} \inf_{u \in S^{\kdim-1}} \|\Amat u\|^2 &
\geq & 1 - \frac{C}{\spgam} \sqrt{\max \left \{
\frac{1}{\kdim}\frac{\log{\inverone}}{\invertwo \log{(p-k)}} , \frac{
\log \invertwo \log(\pdim - \kdim)}{\invertwo \log(\pdim - \kdim)}
\right \}} \\
&=& 1-C T(\gamma, \kdim, \pdim, \invertwo, \inverone)
\end{eqnarray*}

Because $T(\gamma, \kdim, \pdim, \invertwo, \inverone) \rightarrow 0$, for all $\pdim \geq \pdim_1^*$, $\kdim \geq \kdim_1^*$,
$C T(\gamma, \kdim, \pdim, \invertwo, \inverone) < 1$. Thus we can take square root of both sides and apply the identity
$\sqrt{1+x} = 1+\frac{x}{2}+o(x)$ (valid for $|x| < 1$) to conclude that, with probability greater
than $1-\frac{C_1(\pdim_1^*, \kdim_1^*)}{\inverone^2}$:
\begin{eqnarray*}
\frac{1}{\sqrt{{\invertwo \numobs}}} \inf_{u \in S^{\kdim-1}} \|\Amat
u\| & \geq & 1 - \frac{C}{2} T(\gamma, \kdim, \pdim, \invertwo, \inverone) + o(T(\gamma, \kdim, \pdim, \invertwo, \inverone)),
\end{eqnarray*}

As $T(\gamma, \kdim, \pdim, \invertwo, \inverone) \rightarrow 0$, for all $\kdim \geq k_2^*$, $\pdim \geq \pdim_2^*$
 we have that
$| o(T(\gamma, \kdim, \pdim, \invertwo, \inverone))| < \frac{C}{4}T(\gamma, \kdim, \pdim, \invertwo, \inverone)$

Thus, with probability greater
than $1-\frac{C_2(\pdim_1^*, \kdim_1^*, \pdim_2^*, \kdim_2^*)}{\inverone^2}$:
\begin{eqnarray*}
\frac{1}{\sqrt{{\invertwo \numobs}}} \inf_{u \in S^{\kdim-1}} \|\Amat
u\| & \geq & 1 - \frac{3C}{4} T(\gamma, \kdim, \pdim, \invertwo, \inverone),
\end{eqnarray*}

Note that this same process can be repeated to bound the maximum
singular value, yielding the following result:
\begin{eqnarray*}
\frac{1}{\sqrt{{\invertwo \numobs}}} \sup_{u \in S^{\kdim-1}} \|\Amat
 u\| & \leq & 1 + \frac{3C}{4} T(\gamma, \kdim, \pdim, \invertwo, \inverone),
\end{eqnarray*}

Combining these two bounds, we have proved
Lemma~\ref{LemInvertibleSuper}.

\bibliographystyle{plain}

\bibliography{mjwain_super}

\end{document}